\documentclass[letterpaper]{article}
\usepackage{aaai2027}
\nocopyright
\usepackage[hyphens]{url}
\usepackage{graphicx}
\def\UrlFont{\rm}
\usepackage{natbib}
\usepackage{caption}
\usepackage{booktabs}
\usepackage[table]{xcolor}
\usepackage{array}
\usepackage{tabularx}
\usepackage{amsmath}
\usepackage{amssymb}
\usepackage{etoolbox}
\usepackage{multirow}

\definecolor{ResultHighlight}{RGB}{235,235,255}

\definecolor{HeaderFID}{RGB}{232,232,232}
\definecolor{HeaderKID}{RGB}{248,226,228}
\definecolor{HeaderSCA}{RGB}{252,231,204}
\definecolor{HeaderOOB}{RGB}{224,246,220}
\definecolor{HeaderCOL}{RGB}{220,238,248}
\definecolor{HeaderPractical}{RGB}{238,224,243}
\definecolor{VenueText}{RGB}{88,88,88}
\newcommand{\venue}[1]{\,{\textcolor{VenueText}{[#1]}}}

\makeatother
\title{Roomer: Reflective Object-Grounded Model Editing and Repair for 3D Indoor Layout Synthesis}
\author{
    Lingwei Dang\textsuperscript{\rm 1}\equalcontrib, 
    Ziyan Qiu\textsuperscript{\rm 1}\equalcontrib, 
    Jiajia Cheng\textsuperscript{\rm 1}\equalcontrib, 
    Shishuo Shang\textsuperscript{\rm 1}, 
    Zhenhao Zhang\textsuperscript{\rm 2}, \\
    Yufei Zhu\textsuperscript{\rm 2}, 
    Qingxin Xiao\textsuperscript{\rm 1}, 
    Pan Liu\textsuperscript{\rm 1}, 
    Shenghui Huang\textsuperscript{\rm 1}, 
    Yun Hao\textsuperscript{\rm 1}, 
    Juntong Li\textsuperscript{\rm 1}, 
    Qingyao Wu\textsuperscript{\rm 1}\corresponding
}
\affiliations{
    \textsuperscript{\rm 1}School of Software Engineering, South China University of Technology\\
    \textsuperscript{\rm 2}School of Information Science and Technology, ShanghaiTech University \\
    \{levondang, chengjiajia0606\}@163.com, qzy.research@outlook.com, qyw@scut.edu.cn
}

\usepackage{xr}
\begin{document}
\maketitle

\begin{abstract}
Existing indoor layout generators produce globally plausible layouts yet may retain local violations such as collisions, out-of-bounds placements, obstructed openings, and blocked circulation. Most prior work focuses on full-scene synthesis or scene-level optimization, with limited support for identifying responsible objects and locally repairing affected regions. We present Roomer, a reflective repair framework that casts these violations as sparse, object-grounded repair problems. Roomer encodes layouts as ``RoState'' and uses ``RoReview'' to bind measured violations to implicated objects. A geometry-conditioned vision-language model planner proposes a structured local edit, while a deterministic solver validates it and generates a finite set of candidate edits when needed. Each candidate is committed only if full-scene verification confirms that it resolves the target violation without new hard violations or broken protected constraints. We train the planner on Roomer-CC, a controlled-corruption dataset that pairs faulty layouts with object-grounded violation evidence and known-feasible inverse StatePatches. Since existing benchmarks rarely assess whether physically valid layouts are usable, we introduce Roomer-Eval to assess distributional quality, physical validity, and practical usability. Experiments show that Roomer repairs residual violations while preserving valid regions, improves physical validity and usability, and transfers across external generators.
\end{abstract}

\begin{figure}[t]
    \centering
    \includegraphics[width=\columnwidth]{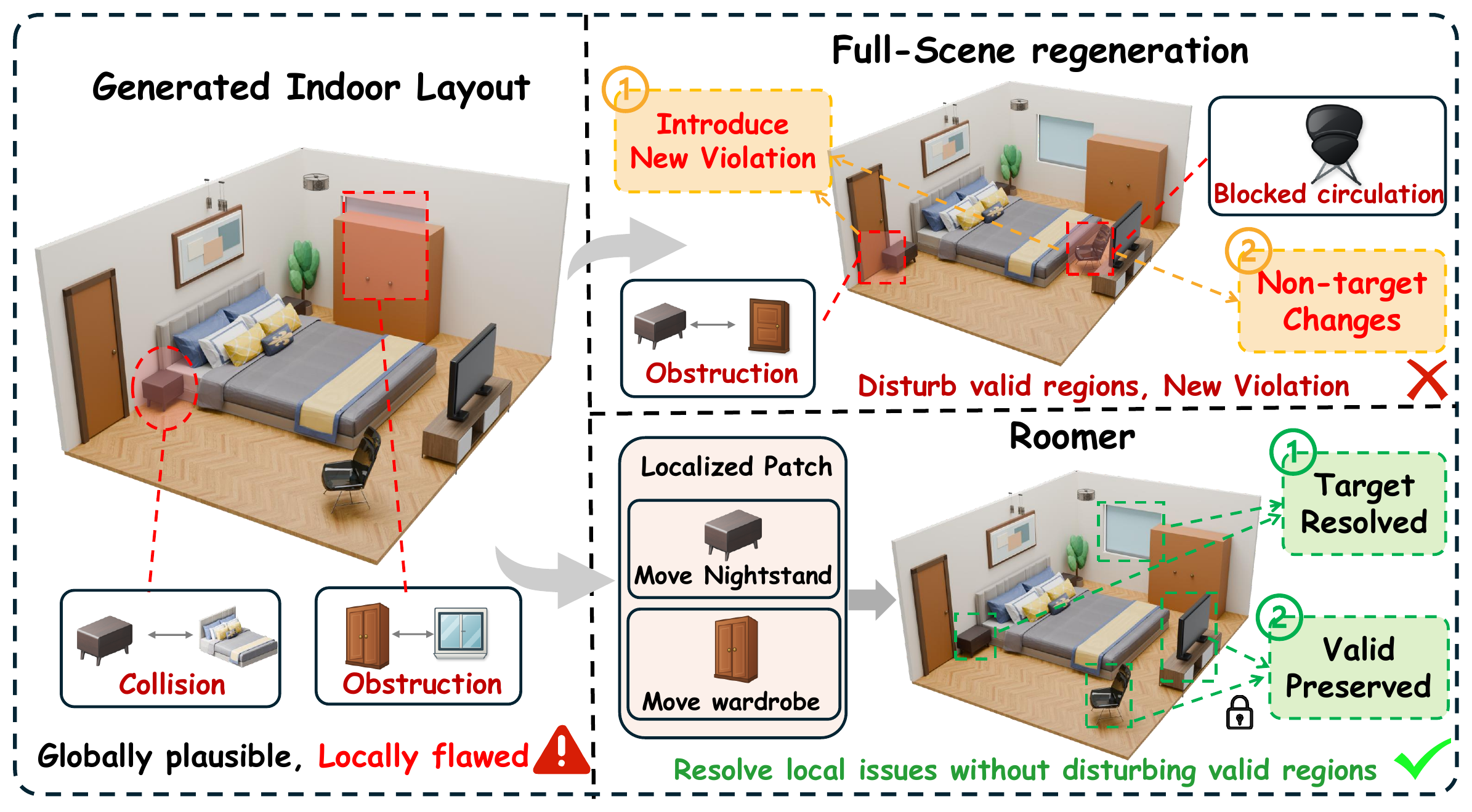}
    \caption{Motivation for Roomer. A generated layout may appear globally plausible yet contain sparse, object-level violations. Full-Scene regeneration may disrupt valid regions and introduce new violations. Roomer instead applies localized, object-grounded StatePatches and commits only patches for which full-scene re-verification confirms target resolution and preservation of already-valid content.}
    \label{fig:teaser}
\end{figure}

\section{Introduction}

Indoor layout synthesis aims to arrange furniture within a given architectural space while satisfying semantic, geometric, and functional requirements. Despite recent progress, existing generators do not always satisfy these requirements. A generated layout may appear visually plausible and semantically coherent while still containing local violations, including collisions, out-of-bounds placements, obstructed architectural openings, and blocked circulation. Correcting these violations through full-scene regeneration may disrupt already-valid furniture configurations and spatial relations. This motivates us to study local repair for indoor layouts: given a completed layout produced by an arbitrary layout generator, the system must identify residual violations, localize the implicated objects, and resolve the violations through a limited set of local edits. The central challenge is to translate measured violation evidence into precise object-level edits and verify that the resulting modifications improve layout quality.

Existing approaches mainly follow two paradigms. Learning-based methods learn furniture composition and spatial distributions from indoor-scene datasets using autoregressive, graph-based, diffusion, and constraint-guided models \cite{paschalidou2021atiss,tang2024diffuscene,lin2024instructscene,yang2024physcene}. Although they generate globally plausible layouts, their scene-level objectives do not explicitly attribute residual violations to responsible objects. LLM-based layout planners instead leverage open-domain knowledge for semantic reasoning and spatial planning, with recent systems incorporating structured scene representations, visual feedback, numerical optimization, and iterative refinement \cite{feng2023layoutgpt,yang2024holodeck,ling2026scenethesis,xia2026sage}. However, these revisions primarily serve prompt satisfaction or scene-level improvement rather than being driven by geometric measurements grounded in specific objects. Both paradigms therefore offer limited support for measurement-grounded object attribution and verifiable repair of residual violations.

Our key insight is that residual violations in generated layouts are typically sparse, localized, and attributable to a small number of objects and their local geometric relations. Such correction should therefore be formulated as verification-gated local state repair rather than full-scene regeneration. Under this formulation, measured evidence is used to identify the responsible objects and guide local edits, while an edit is committed only after full-scene verification confirms that it is both effective and safe.

Based on this insight, we propose Roomer, a reflective repair framework for indoor layouts. Roomer converts a completed layout into RoState, an object-addressable canonical representation, and constructs RoReview to associate each residual violation with geometric measurements and implicated objects. Conditioned on the current layout, RoState, and RoReview, a geometry-conditioned vision-language model planner predicts a schema-constrained StatePatch specifying the repair target, action, and initial parameters. A deterministic solver first validates this proposal and, when necessary, instantiates a finite set of reproducible alternatives. The first candidate that passes full-scene verification is committed; otherwise, the committed layout remains unchanged. Repeating this process yields a controlled and verifiable local repair loop.

Existing indoor-scene datasets do not provide the supervision needed
to train an object-grounded repair planner: they contain complete
layouts but lack paired faulty inputs, attributed violation evidence,
and corrective actions. We therefore construct Roomer-CC by applying
six parameterized object-level corruptions to valid 3D-FRONT layouts,
yielding 67,550 paired repair examples with object-grounded violation
evidence and known-feasible inverse StatePatch targets. These examples
jointly supervise violation attribution and localized repair planning.

Conventional layout benchmarks emphasize distributional similarity and basic physical validity, but rarely assess whether a physically valid layout remains usable. We therefore introduce Roomer-Eval, which combines standard metrics for distributional quality and 3D physical validity
with five reproducible rule families for assessing practical spatial usability. Experiments on 3D-FRONT layouts and outputs from external generators show that Roomer repairs residual violations while preserving valid regions, improving physical validity and practical usability across generators.

Our main contributions are as follows:
\begin{itemize}
    \item We formulate layout correction as
    \textbf{verification-gated local state repair} and introduce Roomer.
    Roomer attributes measured residual violations to implicated objects,
    generates structured local edits, and commits only candidates that pass
    full-scene verification.
    
    \item We construct \textbf{Roomer-CC}, a controlled-corruption
    dataset of 67,550 paired repair examples derived from valid
    3D-FRONT layouts, enabling joint supervision of object-level
    violation attribution and localized repair planning.
    
    \item We introduce \textbf{Roomer-Eval}, a unified evaluation protocol
    that combines standard metrics for distributional quality and 3D physical
    validity with five reproducible rule families for assessing practical
    spatial usability.
\end{itemize}

\section{Related Work}

\begin{figure*}[!t]
    \centering
    \scalebox{1}[0.95]{%
        \includegraphics[width=1\textwidth]{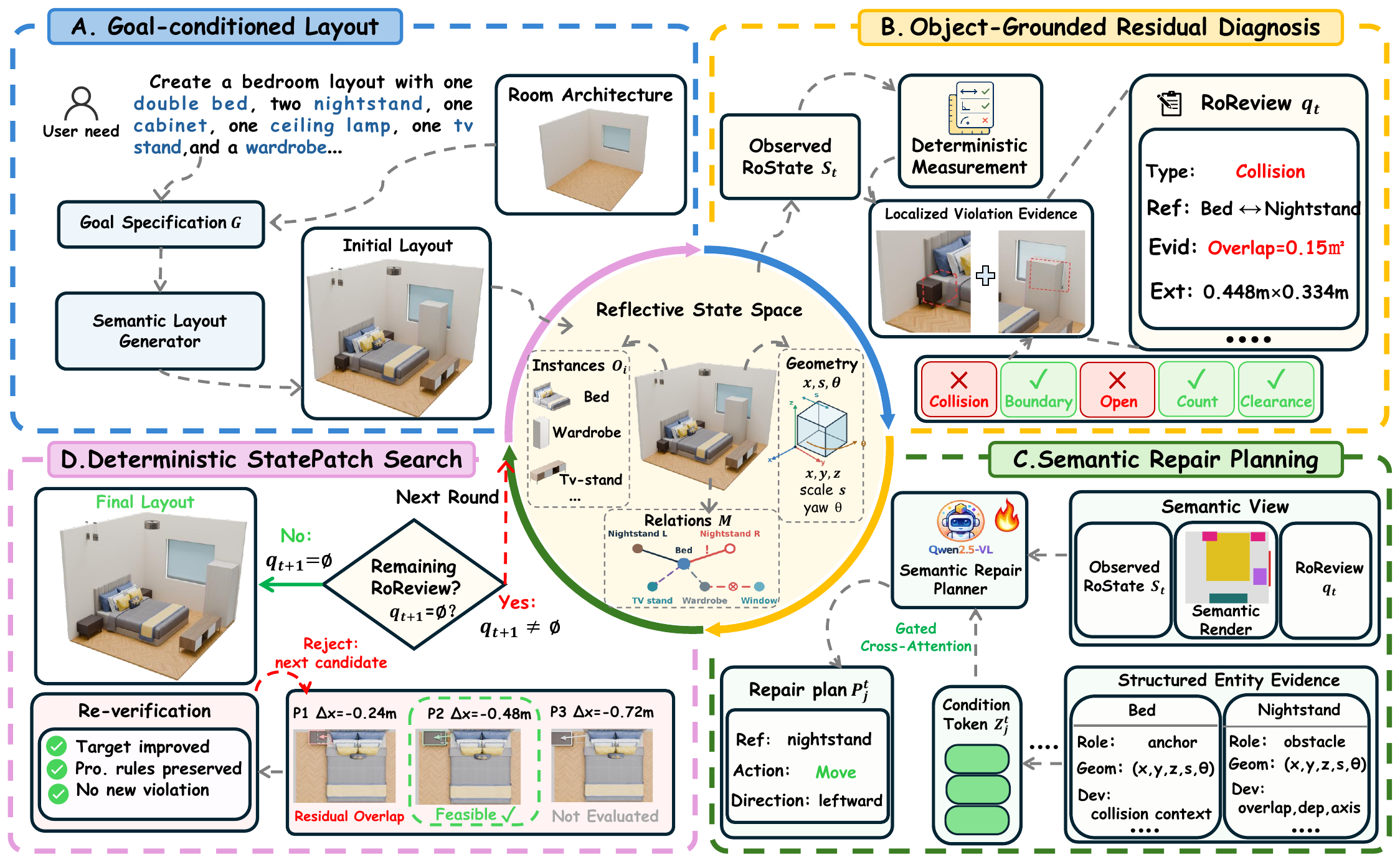}%
    }
    \caption{Overview of Roomer. A generated layout is converted into an object-addressable RoState and evaluated to construct an instance-grounded RoReview. The geometry-conditioned vision-language model planner then predicts an action-specific StatePatch seed; the deterministic solver evaluates the seed first and traverses an ordered fallback sequence only after rejection. The first candidate that passes full-scene verification is committed; otherwise, the transaction is rolled back.}
    \label{fig:framework}
\end{figure*}

\paragraph{Indoor Layout Synthesis.}
Early indoor-layout synthesis methods arranged furniture using explicit
design rules, exemplar statistics, and activity-based priors, often using
search or optimization to obtain feasible configurations under these priors
\cite{yu2011makeithome,merrell2011furniturelayout,
fisher2012objectarrangements,fisher2015activitycentric,
qi2018humancentric}.
Such methods provide explicit control but depend on handcrafted priors and
scene-specific optimization.
Data-driven approaches instead learn complete-layout distributions from
indoor-scene datasets using autoregressive models
\cite{wang2021sceneformer,paschalidou2021atiss},
graph-structured and hierarchical models
\cite{wang2019planit,li2019grains,dhamo2021graph3d,
pei2023scenehgn},
and diffusion-based models
\cite{tang2024diffuscene,lin2024instructscene}.
Constraint-aware variants further incorporate semantic, architectural, or
physical guidance during synthesis to improve relational consistency and
geometric plausibility
\cite{yang2024physcene,sun2026semlayoutdiff}.
More recently, language and vision-language models have been used to
translate open-ended requirements into object lists, spatial relations, and
numerical constraints
\cite{feng2023layoutgpt,yang2024holodeck,anyhome2024,
sun2025hierarchical},
often together with asset retrieval or numerical optimization
\cite{sun2025layoutvlm,casagpt2025,
xiang2026colayout,berdoz2026reason3d}.
These methods broaden semantic controllability and support more open-ended
requirements, but they primarily construct complete scenes or improve
feasibility within the generation pipeline.
Roomer instead targets measurable residual violations in already-generated
layouts, attributes them to implicated object instances, and performs
localized repair under full-scene verification.

\paragraph{Reflective Closed-Loop Scene Editing.}
Feedback-driven methods iteratively evaluate intermediate outputs and use
the resulting feedback to guide subsequent updates. Language-model-based and agentic frameworks rely on self-generated critiques, environmental feedback, or
external tools to refine predictions and actions
\cite{madaan2023selfrefine,shinn2023reflexion,
yao2023react,gou2024critic}. Related closed-loop co-refinement has also been
explored in multimodal human--object interaction synthesis
\cite{dang2025svimo,dang2026harmohoi}. For explicit 3D scenes, prior methods
broadly follow two directions. Instruction-driven editors translate
user-specified goals into compositional object operations or physically
valid action sequences
\cite{zheng2025editroom,bucher2026respace,
noh2026editasact}, whereas scene-feedback-driven systems revise generated
scenes over multiple rounds using semantic, visual, geometric, physical, or
functional assessments
\cite{yang2025sceneweaver,ling2026scenethesis,
xia2026sage,zhao2026scenerevis,wang2026function2scene}. These studies
demonstrate the value of iterative correction, but their objectives are
typically specified by user instructions, model-generated critiques, or
scene-level quality signals rather than deterministic measurements
attributed to specific object instances. Roomer instead derives repair
targets from measured residual violations, attributes them to implicated
objects, and commits only localized edits that pass full-scene verification without regressions.

\paragraph{Physical Validity and Practical Spatial Usability.}
Indoor layout evaluation has primarily focused on distributional quality
and basic physical validity. FID, KID, generative precision and recall,
and category statistics measure similarity to real-scene distributions
and furniture compositions, whereas collision and out-of-bounds rates
assess geometric feasibility
\cite{heusel2017fid,binkowski2018kid,
kynkaanniemi2019precisionrecall,paschalidou2021atiss,
tang2024diffuscene,yang2024physcene}.
SceneEval \cite{tam2026sceneeval} further assesses compliance with explicit object and relation requirements and evaluates support, collision, and navigability.
However, human-centered and ergonomics-oriented studies show that layouts
may remain unusable even when collision-free and within room boundaries
if they lack sufficient clearance for approach, interaction, or circulation
\cite{fisher2015activitycentric,qi2018humancentric,
leimer2022layoutenhancer}.
Practical spatial usability should therefore be evaluated as a distinct dimension rather than inferred from distributional quality or physical validity. Roomer-Eval complements existing metrics with deterministic and reproducible rules for functional organization, operational clearance,
and circulation in residential layouts.

\section{Method}

\subsection{Problem Formulation and Overview}
Roomer repairs residual violations in a complete indoor layout without requiring a user-specified edit target or regenerating the entire scene. Given an architectural environment $\mathcal{A}$, a target room specification $\mathcal{G}$ containing textual user requirements and structured room constraints, and an initial layout $\mathcal{Y}^{0}$ produced by an arbitrary upstream generator, Roomer repeatedly (i) attributes one detected violation to the implicated scene entities, (ii) predicts a local StatePatch, and (iii) commits an instantiated candidate only after full-scene verification.

Let $t\in\{0,\ldots,K-1\}$ index planning attempts and let $p^{t,\star}$ denote the accepted candidate, if one exists. The committed layout evolves as
\begin{equation}
\mathcal{Y}^{t+1}
=
\begin{cases}
\mathcal{Y}^{t}\oplus p^{t,\star}, & p^{t,\star}\neq\varnothing,\\
\mathcal{Y}^{t}, & p^{t,\star}=\varnothing,
\end{cases}
\label{eq:roomer-transition}
\end{equation}
where $\oplus$ applies an object-level state update. Rejected attempts consume the planning budget but never modify the committed state. Roomer is generator-agnostic and requires only that the upstream output be convertible to the canonical representation summarized in Fig.~\ref{fig:framework}.

\paragraph{Initial layout generation.}
We use a Qwen-Image-based upstream generator, fine-tuned to map an
architectural condition map and structured room prompt to a fixed-palette
top-down semantic layout. A parser converts this output into
the structured initial layout $\mathcal{Y}^{0}$ by mapping RGB values to
semantic labels, filtering artifacts and room-incompatible categories,
extracting connected components as furniture instances, and augmenting
their planar geometry with attributes predicted by a pretrained model.

\subsection{Instance-Grounded Repair Context}
At attempt $t$, Roomer represents the current layout as the object-addressable RoState $\mathcal{S}^{t}=(\mathcal{A},\mathcal{Y}^{t},\mathcal{F}^{t},\Pi^{t})$, where $\mathcal{F}^{t}$ contains geometry-derived functional regions and $\Pi^{t}$ maps stable references to scene entities. The rule evaluator detects applicable rule instances, and the scheduler selects an active instance $v^{t}=(r^{t},\rho^{t})$, comprising a rule family and the stable references to its participating entities.

The active instance is serialized into the following RoReview tuple: $q^{t}=(\kappa(v^{t}),\tau^{t},\mathcal{E}^{t},\mathcal{W}^{t},\boldsymbol{\delta}^{t})$.
Its fields encode the stable issue key, violation type, participating entities, relational roles, and type-specific measurements. For example, a bedside-clearance review identifies the bed as the anchor, the clearance region as protected, and intruding furniture as obstructors, together with the clearance deficit and intrusion ratio. The same representation supports collision, out-of-bounds, opening-obstruction, and passage-width violations. RoReview thus converts an abstract rule failure into measurable, attributable evidence without prescribing the repair action. Implementation details are provided in Supplementary Sec.~\ref{app:state-review}.

\subsection{Geometry-Conditioned StatePatch Planning}
For the active RoReview entry $q^{t}$, the planner receives the current top-down semantic rendering $I^{t}$ of $\mathcal{Y}^{t}$, the target room specification $\mathcal{G}$, and serialized RoState and RoReview. Although these inputs identify the relevant entities and roles, text serialization may obscure the metric differences in distance, direction, overlap, and clearance that determine local feasibility. We therefore align structured geometric evidence with the semantic context through stable references.

Each participating entity is encoded by its category, role, editability, normalized floor-plane geometry, and type-specific measurements. Learnable Repair Queries aggregate the masked evidence as $\mathbf{Z}^{t}=E_{\phi}(\mathbf{Q},\mathbf{X}^{t};\mathbf{M}^{t})$, where $\mathbf{M}^{t}$ masks padded entities. Let $\bar{\mathbf{H}}_{\ell,\mathrm{sa}}^{t}=\operatorname{Norm}(\mathbf{H}_{\ell,\mathrm{sa}}^{t})$ and $\bar{\mathbf{Z}}^{t}=\operatorname{Norm}(\mathbf{Z}^{t})$ denote the normalized language and geometry states, respectively. At selected decoder layers, the geometry tokens condition the language states through a residual cross-attention adapter:
\begin{equation}
\widetilde{\mathbf{H}}_{\ell}^{t}
=
\mathbf{H}_{\ell,\mathrm{sa}}^{t}
+
\gamma_{\ell}\mathbf{W}_{\ell}^{o}
\operatorname{CrossAttn}_{\ell}\!\left(
\mathbf{W}_{\ell}^{q}\bar{\mathbf{H}}_{\ell,\mathrm{sa}}^{t},
\bar{\mathbf{Z}}^{t},
\bar{\mathbf{Z}}^{t}
\right).
\label{eq:geometry-condition-injection}
\end{equation}
Implementation details are provided in Supplementary Sec.~\ref{app:geometry-conditioning}. The planner then predicts a schema-constrained StatePatch with action, target, and parameter-seed fields, $\widehat{p}^{t}=(\widehat{a}^{t},\allowbreak\widehat{\boldsymbol{\xi}}^{t},\allowbreak\widehat{\boldsymbol{\alpha}}_{0}^{t})$. The action field $\widehat{a}^{t}$ selects one of six operations: \texttt{MOVE}, \texttt{ROTATE}, \texttt{SCALE}, \texttt{INSERT}, \texttt{DELETE}, or \texttt{REPLACE}. The target field $\widehat{\boldsymbol{\xi}}^{t}$ specifies the action-dependent target, while $\widehat{\boldsymbol{\alpha}}_{0}^{t}$ provides its parameter seed. The planner determines what to edit and which action to take, while the deterministic solver realizes precise candidates; the seed is never committed directly. Representative before-and-after examples of all six actions are shown in Supplementary Fig.~\ref{fig:statepatch-action-examples}.

\begin{table*}[!t]
\centering
\small
\setlength{\tabcolsep}{3.2pt}
\begin{tabularx}{\textwidth}{@{}>{\raggedright\arraybackslash}p{0.32\textwidth}*{6}{>{\centering\arraybackslash}X}@{}}
\toprule
\textbf{Method}
& \cellcolor{HeaderFID}\textbf{FID}$\downarrow$
& \cellcolor{HeaderKID}\textbf{KID} $\times 10^3\!\downarrow$
& \cellcolor{HeaderSCA}\textbf{SCA Gap}$\downarrow$
& \cellcolor{HeaderOOB}\textbf{OOB}$\downarrow$
& \cellcolor{HeaderCOL}\textbf{COL}$\downarrow$
& \cellcolor{HeaderPractical}\textbf{Practical}$\uparrow$ \\
\midrule
DiffuScene-RS\venue{CVPR 2024} & 68.15 & 11.81 & 16.57 & 38.18\% & 28.44\% & 45.09\% \\
InstructScene\venue{ICLR 2024} & 64.90 & 7.21 & 19.98 & 37.49\% & 28.50\% & 40.99\% \\
SemLayoutDiff-RS\venue{3DV 2026} & 90.68 & 29.68 & 34.12 & 49.43\% & 60.42\% & 57.77\% \\
ReSpace\venue{ES-Reasoning @ ICLR 2026} & \textbf{57.42} & \textbf{5.10} & 21.73 & 14.72\% & 36.56\% & 66.43\% \\
Ours-Initial & 64.50 & 10.80 & 19.20 & 24.40\% & 29.52\% & 72.50\% \\
\rowcolor{ResultHighlight}
\textbf{Ours-Final} & 60.20 & 7.82 & \textbf{12.46} & \textbf{8.66\%} & \textbf{17.50\%} & \textbf{82.98\%} \\
\bottomrule
\end{tabularx}
\caption{Density-controlled comparison on \textit{common-1100}. FID, KID, and SCA Gap are unweighted Macro-3 averages; OOB, COL, and Practical use all 1,100 scenes. Best results are bolded.}
\label{tab:main-results}
\end{table*}

\subsection{Deterministic Instantiation and Verification-Gated Commit}
The planner determines the repair target and action, whereas the deterministic solver resolves numerical uncertainty. After schema, reference, and action validation, the solver evaluates the normalized planner seed first and, only after rejection, instantiates an action-specific neighborhood ordered by seed distance and edit magnitude. Evaluation stops at the first candidate that passes the verification gate; otherwise, the solver returns $\varnothing$. Exact domains and fallback sequences are provided in Supplementary Sec.~\ref{app:solver-control}.

Each candidate $p$ is applied only to a temporary layout $\mathcal{Y}_{p}^{t}\triangleq\mathcal{Y}^{t}\oplus p$, after which all functional regions and rule instances are recomputed. Let $\mathcal{H}$ be the stable hard-violation key set, $\mathcal{Z}_{\mathrm{keep}}^{t}$ the protected satisfied relations, $\mathcal{C}$ the structural-validity predicate, and $\mathcal{V}$ the family-balanced residual over hard, content, relational, and practical repair rules. For compactness, write $\mathcal{F}_{p}^{t}=\mathcal{F}(\mathcal{Y}_{p}^{t};\mathcal{A},\mathcal{G})$ and $\mathcal{F}^{t}=\mathcal{F}(\mathcal{Y}^{t};\mathcal{A},\mathcal{G})$ for $\mathcal{F}\in\{\mathcal{H},\mathcal{Z},\mathcal{C},\mathcal{V}\}$. The candidate is accepted if and only if
\begin{equation}
\mathsf{Acc}_{t}(p)=1
\Longleftrightarrow
\left\{
\begin{aligned}
&\mathsf{Target}_{t}(p)=1,\\
&\mathcal{H}_{p}^{t}\setminus\mathcal{H}^{t}=\varnothing,\\
&\mathcal{Z}_{\mathrm{keep}}^{t}\subseteq\mathcal{Z}_{p}^{t},\\
&\mathcal{C}_{p}^{t}=1,\\
&\mathcal{V}_{p}^{t}\leq\mathcal{V}^{t}-\varepsilon_V.
\end{aligned}
\right.
\label{eq:verification-gate}
\end{equation}
These conditions enforce target effectiveness, physical safety, relation preservation, structural validity, and global progress. Thus, reducing the active violation is insufficient unless the complete scene remains valid. Definitions and thresholds are provided in Supplementary Sec.~\ref{app:verification}. Issues without an acceptable candidate are blocked in the current state; any successful commit clears the blocked set because the geometry has changed. The loop stops when no violations remain, all detected issues are blocked, or the budget $K$ is exhausted.

\subsection{Controlled-Corruption Supervision}
Roomer-CC derives object-grounded repair supervision from valid
3D-FRONT layouts that satisfy their target room specifications
\cite{fu2021threedfront}. For each example, we apply one of six
parameterized object-level corruptions and recompute the semantic
observation, RoState, RoReview, and geometric evidence. The corrupted
layout serves as the repair input, while the known inverse corruption
defines a feasible StatePatch target that restores the valid reference
layout. This pairing associates each measured violation and its
implicated object or missing role with an executable correction.

We retain a pair only when the intended violation is triggered,
attributed to the expected object or role, and removed by the inverse
StatePatch without introducing a new hard violation or breaking a
protected relation. Construction, splits, and action--issue coverage
are provided in Supplementary Secs.~\ref{app:data-partitions}
and~\ref{app:solver-control}.

Continuous parameters are serialized in the StatePatch and optimized with a field-weighted next-token objective that emphasizes action and target tokens; serialization and weighting details are provided in Supplementary Sec.~\ref{app:numeric-ar}.

\section{Roomer-Eval: Unified Evaluation Protocol}
\label{sec:roomer-eval}

Roomer-Eval combines standard distributional and physical metrics with
Practical, our rule-based measure of spatial usability. FID, KID, and
SCA Gap are computed on shared semantic renderings; out-of-bounds
placement (OOB) and mesh-level collision (COL) are computed on assembled
scenes. Practical captures functional failures that can persist even when
a layout is collision-free and within the room boundary.

For rule instance $j$ in scene $i$, let
$a_{ij},s_{ij}\in\{0,1\}$ denote applicability and satisfaction.
Practical is the micro-average over applicable instances:
\begin{equation}
\operatorname{Practical}
=
\frac{\sum_{i,j} a_{ij}s_{ij}}
     {\sum_{i,j} a_{ij}}.
\label{eq:practical-usability}
\end{equation}
Thus, each applicable rule instance contributes equally, while N/A
instances are excluded. Object-level families may yield multiple
instances per scene, whereas room-level families yield at most one.

The five rule families cover living-room functional organization,
dining-table clearance, door swing-proxy avoidance, walkable connectivity,
and bedside clearance. Complete applicability, association, exemption,
geometry, and contact definitions are provided in Supplementary
Sec.~\ref{app:roomer-eval-rules}.

Although the same frozen definitions guide Roomer during repair, final
scores are recomputed from each method's complete output without reusing
repair-time detections or decisions. Practical measures instance-level
rule compliance rather than scene-level all-pass performance, building-code
compliance, exhaustive ergonomics, or aesthetic quality.

\section{Experiments}
\label{sec:experiments}

Unless otherwise stated, all quantitative experiments use the frozen \textit{common-1100} cohort with a shared evaluator and at most ten repair rounds; professional validation uses a separate frozen sample.

\subsection{Experimental Setup}
\label{sec:experimental-setup}

\paragraph{Data and baselines.}
The frozen \textit{common-1100} cohort contains Qwen-Image outputs \cite{wu2025qwenimage} for 1,100 held-out 3D-FRONT/3D-FUTURE rooms \cite{fu2021threedfront,fu2021threedfuture}: 777 bedrooms, 155 living rooms, and 168 dining rooms. It is excluded from all training and model selection. Roomer-CC contains 61,010 training, 3,270 validation, and 3,270 held-out test examples. We compare with ReSpace \cite{bucher2026respace}, DiffuScene \cite{tang2024diffuscene}, InstructScene \cite{lin2024instructscene}, and SemLayoutDiff \cite{sun2026semlayoutdiff}, retaining each method's native conditioning, asset-retrieval, and assembly pipeline. Density-controlled selection uses only output validity and floor-standing object count, never evaluation metrics. Complete split provenance, baseline interfaces, and selection rules are provided in the supplementary material.

\paragraph{Implementation.}
Ours-Initial is generated at $512\times512$ resolution using
Qwen-Image~\cite{wu2025qwenimage} with rank-64 DiT LoRA.
The repair planner is initialized from
Qwen2.5-VL-7B-Instruct~\cite{bai2025qwen25vl}, with the
backbone frozen and rank-8 LoRA adapters and
geometry-conditioning modules optimized. The geometry branch
encodes up to 16 evidence entities as 512-dimensional tokens,
aggregates them with eight Repair Queries, and injects residual
cross-attention into decoder layers 22, 24, 26, and 28.

Qwen-Image is trained for four epochs with AdamW on one
NVIDIA A800 80GB GPU. The planner is trained for five epochs
on two A800 GPUs using learning rates of $1e-5$ and
$1e-4$ for the LoRA and geometry parameters, respectively,
cosine scheduling with 5\% warmup, and an effective batch size
of 32. Full optimization details are provided in the
supplementary material.

\begin{figure*}[!t]
    \centering
    \scalebox{1}[0.96]{%
        \includegraphics[width=\textwidth]{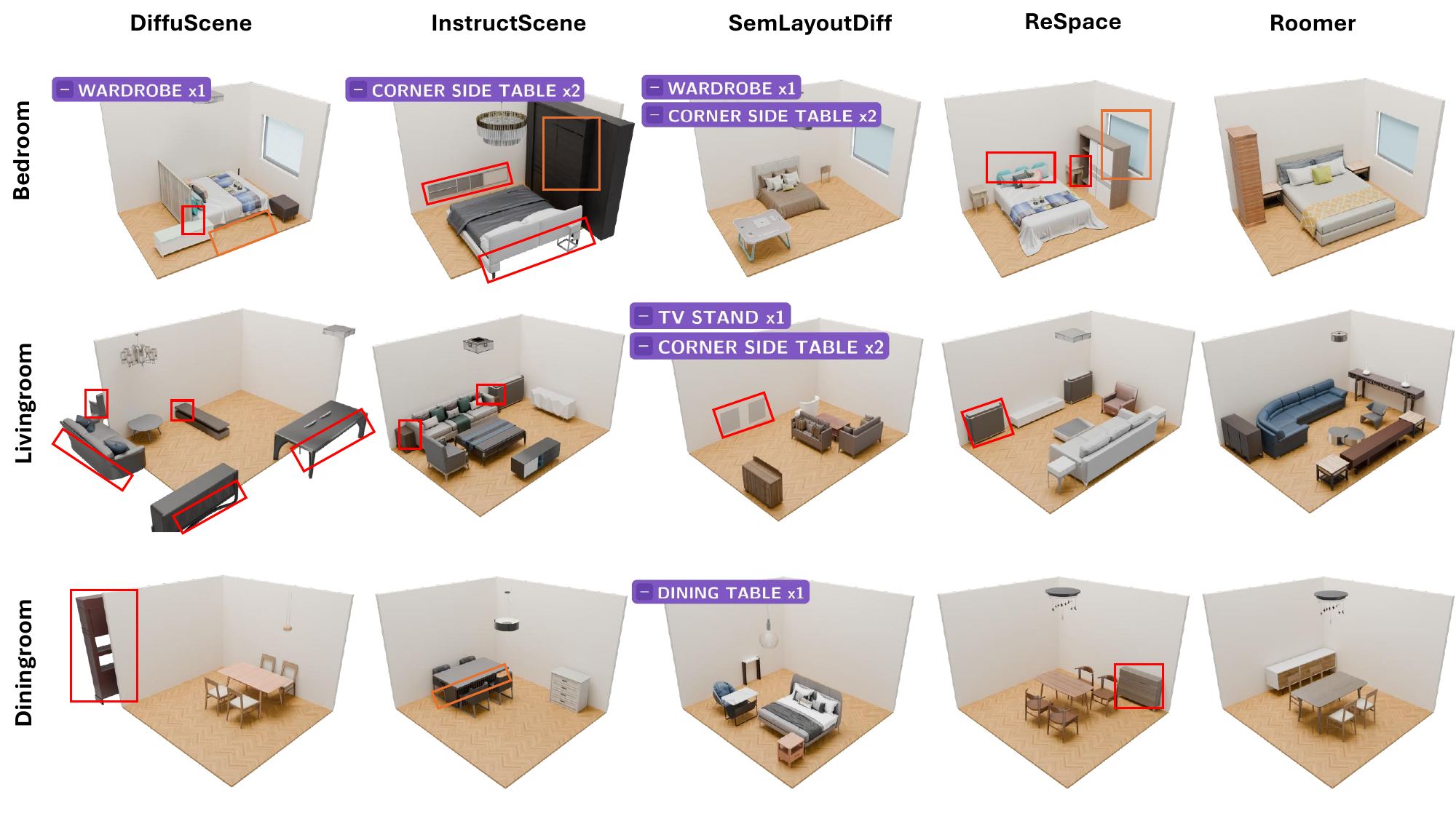}%
    }
    \caption{Representative final-layout comparisons on
    \textit{common-1100}. Red boxes highlight geometric violations, orange boxes indicate practical violations, and purple tags denote unmet furniture requirements
    caused by missing or mismatched objects.}
    \label{fig:baseline-qualitative}
\end{figure*}

\subsection{Density-Controlled Final-Layout Comparison}
\label{sec:final-layout-comparison}

Table~\ref{tab:main-results} compares density-controlled complete outputs under a shared renderer and evaluator. Ours-Final achieves the best SCA Gap, OOB, COL, and Practical, while ReSpace remains strongest on FID and KID. Relative to Ours-Initial, Roomer reduces OOB and COL by 15.74 and 12.02 percentage points and raises Practical by 10.48 points; all three distributional metrics also improve. This pattern shows that local repair primarily strengthens physical validity and practical usability without degrading distributional quality relative to its own initialization, although ReSpace retains an advantage in FID and KID. Figure~\ref{fig:baseline-qualitative} complements the aggregate results: the baselines exhibit unmet furniture requirements and localized geometric or practical violations, whereas Roomer satisfies the required furniture content and avoids the highlighted failures
across all three room types.

\subsection{Repairing Outputs from External Generators}
\label{sec:generator-transfer}

To evaluate the applicability of Roomer to layouts produced by other
generators, we apply the same frozen repair procedure to outputs from
four external generators, retaining all scenes and rolling back rejected
candidates. Table~\ref{tab:generator-transfer} shows that OOB, COL, and
Practical improve for every generator, demonstrating that Roomer can
effectively repair layouts with different error patterns. Distributional
metrics also improve in 10 of 12 pairs; ReSpace KID and DiffuScene-RS
SCA Gap are the only exceptions, with complete values reported in
Supplementary Table~\ref{tab:generator-transfer-distribution}. The
remaining errors vary across generators, with SemLayoutDiff-RS retaining
higher OOB and COL after repair. Overall, Roomer consistently improves
outputs from multiple external generators, although the final quality
remains influenced by the upstream layouts.

\begin{table}[!t]
\centering
\small
\begin{tabular}{@{}lcrrr@{}}
\toprule
\textbf{Method} & \textbf{State} & \textbf{OOB}$\downarrow$ & \textbf{COL}$\downarrow$ & \textbf{Practical}$\uparrow$ \\
\midrule
\multirow{2}{*}{ReSpace}
& Before & 14.72 & 36.56 & 66.43 \\
& \cellcolor{ResultHighlight}\textbf{After}
& \cellcolor{ResultHighlight}\textbf{1.16}
& \cellcolor{ResultHighlight}\textbf{7.61}
& \cellcolor{ResultHighlight}\textbf{75.12} \\
\addlinespace[1pt]
\multirow{2}{*}{DiffuScene-RS}
& Before & 38.18 & 28.44 & 45.09 \\
& \cellcolor{ResultHighlight}\textbf{After}
& \cellcolor{ResultHighlight}\textbf{6.98}
& \cellcolor{ResultHighlight}\textbf{6.95}
& \cellcolor{ResultHighlight}\textbf{47.73} \\
\addlinespace[1pt]
\multirow{2}{*}{InstructScene}
& Before & 37.49 & 28.50 & 40.99 \\
& \cellcolor{ResultHighlight}\textbf{After}
& \cellcolor{ResultHighlight}\textbf{5.67}
& \cellcolor{ResultHighlight}\textbf{5.88}
& \cellcolor{ResultHighlight}\textbf{45.70} \\
\addlinespace[1pt]
\multirow{2}{*}{SemLayoutDiff-RS}
& Before & 49.43 & 60.42 & 57.77 \\
& \cellcolor{ResultHighlight}\textbf{After}
& \cellcolor{ResultHighlight}\textbf{32.83}
& \cellcolor{ResultHighlight}\textbf{43.92}
& \cellcolor{ResultHighlight}\textbf{66.86} \\
\bottomrule
\end{tabular}
\caption{Paired transfer results on frozen outputs from four external generators; all 1,100 scenes are retained. Bold marks the better result within each Before--After pair.}
\label{tab:generator-transfer}
\end{table}

\newpage

\begin{figure*}[!t]
\centering
\includegraphics[width=0.97\textwidth]{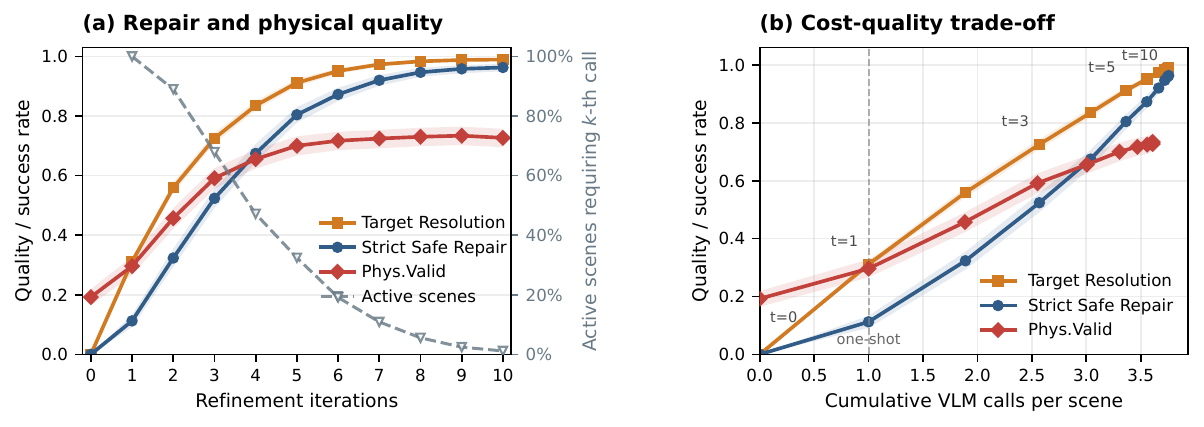}
\caption{Round-wise convergence and cost--quality trade-off on \textit{common-1100}. Adaptive calls concentrate on a shrinking set of unresolved scenes.}
\label{fig:component-ablation-curves}
\end{figure*}

\subsection{Controlled Analysis of the Roomer Repair Loop}

\label{sec:controlled-roomer-loop}

\paragraph{Protocol and metrics.}
All variants start from byte-identical layouts and RoReview issues and share the same action space, round budget, and stopping conditions. Hard, Prac., Tgt., New, NT, and Safe denote hard validity, Practical, target resolution, new hard errors, non-target preservation, and strict safe repair, respectively; RB denotes rollback. Full metric definitions and round-prefix results are provided in Supplementary Secs.~\ref{app:controlled-metrics} and~\ref{app:additional-ablations}.

\begin{table}[!t]
\centering
\small
\setlength{\tabcolsep}{1mm}
\begin{tabular}{@{}lcccccc@{}}
\toprule
\textbf{Variant}
& \textbf{Hard}$\uparrow$
& \textbf{Prac.}$\uparrow$
& \textbf{Tgt.}$\uparrow$
& \textbf{New}$\downarrow$
& \textbf{NT}$\uparrow$
& \textbf{Safe}$\uparrow$ \\
\midrule
Ours-Init. & 19.27 & 72.50 & -- & -- & -- & -- \\
w/o RoReview & 19.73 & 50.39 & 2.44 & 0.00 & 94.15 & 0.45 \\
w/o Geom. Adapt. & 62.82 & 74.90 & 88.40 & 0.00 & 99.81 & 81.64 \\
Direct Seed & 48.00 & 67.80 & 65.70 & 0.00 & 99.93 & 55.18 \\
Seed-free & 68.82 & 77.60 & 90.70 & 0.00 & 99.91 & 89.64 \\
Full JSON & 16.55 & 49.82 & 32.48 & 32.64 & 48.61 & 1.91 \\
w/o Verif./RB & 23.55 & 71.72 & 64.91 & 54.27 & 97.63 & 23.18 \\
\rowcolor{ResultHighlight}
\textbf{Ours-Final} & \textbf{72.64} & \textbf{82.98} & \textbf{98.96} & \textbf{0.00} & \textbf{99.94} & \textbf{96.27} \\
\bottomrule
\end{tabular}
\caption{Controlled ablation on identical \textit{common-1100} initial states. Ours-Init./Final denote $K=0/10$; all values are percentages.}
\label{tab:component-ablation}
\end{table}
\paragraph{Component ablation.}
Table~\ref{tab:component-ablation} isolates object-grounded diagnosis, geometry conditioning, candidate generation, and safe commitment. Without RoReview, Target Resolution falls to 2.44\% and Practical drops below the initial layout, showing that identifying the responsible object is a prerequisite rather than an auxiliary cue. Removing geometry conditioning causes a smaller but substantial degradation, indicating that measured geometry mainly improves execution once the repair target is known. The candidate variants further show that deterministic search supplies most of the robustness, while the planner seed provides complementary coverage. Full-Scene JSON Rewrite and removing verification expose different failure modes: rewriting the complete scene breaks locality and non-target preservation, whereas removing verification retains mostly local edits but commits many new hard errors. The complete method avoids both failure modes, supporting the joint need for structured local StatePatches and verification-gated commitment.

\paragraph{Iterative benefit and adaptive computation.}

To isolate later-round benefits, we evaluate trajectory prefixes at $K\in\{1,3,5,7,10\}$ from the same frozen $K=10$ runs rather than resampling. Figure~\ref{fig:component-ablation-curves} shows that most gains occur in the early rounds, while later rounds resolve a shrinking tail of difficult scenes and continue to improve safe repair. Because resolved scenes leave the active set, Roomer averages 3.756 planner calls per scene, 62.44\% fewer than fixed ten-round execution. Adaptive stopping therefore preserves the long-tail benefit of iterative repair without applying the full budget uniformly to every scene.

\subsection{Professional Validation of Practical Usability}
\label{sec:professional-validation}

To test whether Practical aligns with professional usability
judgments, ten evaluators with interior-design experience assess
an independent frozen set of 90 layouts sampled from the evaluated
baselines and disjoint from the Roomer repair cohort. We sample
30 layouts per Practical level, balance room types, randomize
presentation, and blind evaluators to generator identity and
Practical score.
Each evaluator rates all layouts, yielding 900 binary judgments;
approval requires at least seven ``Yes'' votes. Full protocol
details are provided in the supplementary material.

\begin{table}[!t]
\centering
\small
\setlength{\tabcolsep}{1mm}
\begin{tabular}{@{}
>{\raggedright\arraybackslash}p{0.300\columnwidth}
>{\centering\arraybackslash}p{0.160\columnwidth}
>{\centering\arraybackslash}p{0.140\columnwidth}
>{\centering\arraybackslash}p{0.330\columnwidth}
@{}}
\toprule
\textbf{Practical Level} & \textbf{Approved} &
\textbf{Rate} & \textbf{95\% CI} \\
\midrule
Low    & 7/30  & 23.3\% & [11.8\%, 40.9\%] \\
Medium & 17/30 & 56.7\% & [39.2\%, 72.6\%] \\
High   & 25/30 & 83.3\% & [66.4\%, 92.7\%] \\
\bottomrule
\end{tabular}
\caption{Professional approval across Practical levels with
Wilson 95\% confidence intervals.}
\label{tab:professional-validation}
\end{table}

Table~\ref{tab:professional-validation} shows that professional
approval increases monotonically from Low to High Practical, with a
significant ordered trend ($Z=4.67$, $p=3.07\times10^{-6}$). This
strong association supports Practical as a meaningful indicator of
professionally judged spatial usability. Accordingly, the higher
Practical scores achieved by Roomer reflect improvements that are
aligned with professional usability judgments.

\section{Conclusions}
We presented Roomer, a reflective local repair framework that formulates
residual failures in 3D indoor layouts as measurable, object-grounded
repair tasks. It combines RoReview-based diagnosis,
geometry-conditioned StatePatch planning, deterministic candidate
search, and verification-gated commitment. Roomer-CC provides
object-grounded repair supervision, while Roomer-Eval jointly assesses
distributional quality, physical validity, and practical spatial
usability. Experiments on frozen layouts and four external generators
show consistent gains in physical validity and Practical, whose
agreement with professional judgments is independently supported.
These results show that reliable local correction requires both
object-grounded violation attribution and verification of candidate
state transitions, rather than regenerating already-valid content.
Roomer is currently limited to violations covered by its predefined
residential rule set and to repairs reachable within its finite
action-specific candidate spaces; issues outside this coverage or
without an acceptable candidate remain unresolved. Future work will extend the rule library and candidate generation to support broader room types and a wider range of functional
requirements.

\bibliography{loreflection_references}

\clearpage
\def\StandaloneSupp{}
\ifdefined\StandaloneSupp\else
\documentclass[letterpaper]{article}
\usepackage{aaai2027}  
\nocopyright
\usepackage[hyphens]{url}
\usepackage{graphicx}
\urlstyle{rm}
\def\UrlFont{\rm}
\usepackage{natbib}
\usepackage{caption}
\usepackage{booktabs}
\usepackage[table]{xcolor}
\usepackage{array}
\usepackage{tabularx}
\usepackage{amsmath}
\usepackage{amssymb}
\usepackage{etoolbox}
\usepackage{multirow}

\definecolor{ResultHighlight}{RGB}{235,235,255}
\definecolor{HeaderFID}{RGB}{232,232,232}
\definecolor{HeaderKID}{RGB}{248,226,228}
\definecolor{HeaderSCA}{RGB}{252,231,204}
\definecolor{HeaderOOB}{RGB}{224,246,220}
\definecolor{HeaderCOL}{RGB}{220,238,248}
\definecolor{HeaderPractical}{RGB}{238,224,243}
\definecolor{VenueText}{RGB}{88,88,88}
\newcommand{\venue}[1]{\,{\textcolor{VenueText}{[#1]}}}
\newcommand{\vheadl}[1]{\raisebox{-.5\height}{\shortstack[l]{#1}}}
\newcommand{\vheadc}[1]{\raisebox{-.5\height}{\shortstack[c]{#1}}}

\frenchspacing
\pdfinfo{/TemplateVersion (2027.1)}
\setcounter{secnumdepth}{0}

\title{Roomer: Reflective Object-Grounded Model Editing and Repair for 3D Indoor Layout Synthesis\\Supplementary Material}

\begin{document}
\maketitle
\fi

\ifdefined\StandaloneSupp
\twocolumn[{%
\begin{center}
{\Large\bfseries Roomer: Reflective Object-Grounded Model Editing and Repair for \\ 3D Indoor Layout Synthesis\par}
\vspace{0.35em}
{\large Supplementary Material\par}
\vspace{0.6em}
\end{center}
\vspace{0.8em}
}]
\fi

\providecommand{\vheadl}[1]{\raisebox{-.5\height}{\shortstack[l]{#1}}}
\providecommand{\vheadc}[1]{\raisebox{-.5\height}{\shortstack[c]{#1}}}

\appendix
\setcounter{secnumdepth}{1}
\setcounter{table}{0}
\setcounter{figure}{0}
\setcounter{equation}{0}
\renewcommand{\thetable}{A\arabic{table}}
\renewcommand{\thefigure}{A\arabic{figure}}
\renewcommand{\theequation}{A\arabic{equation}}

\begin{figure*}[!t]
    \centering
    \includegraphics[width=\textwidth]{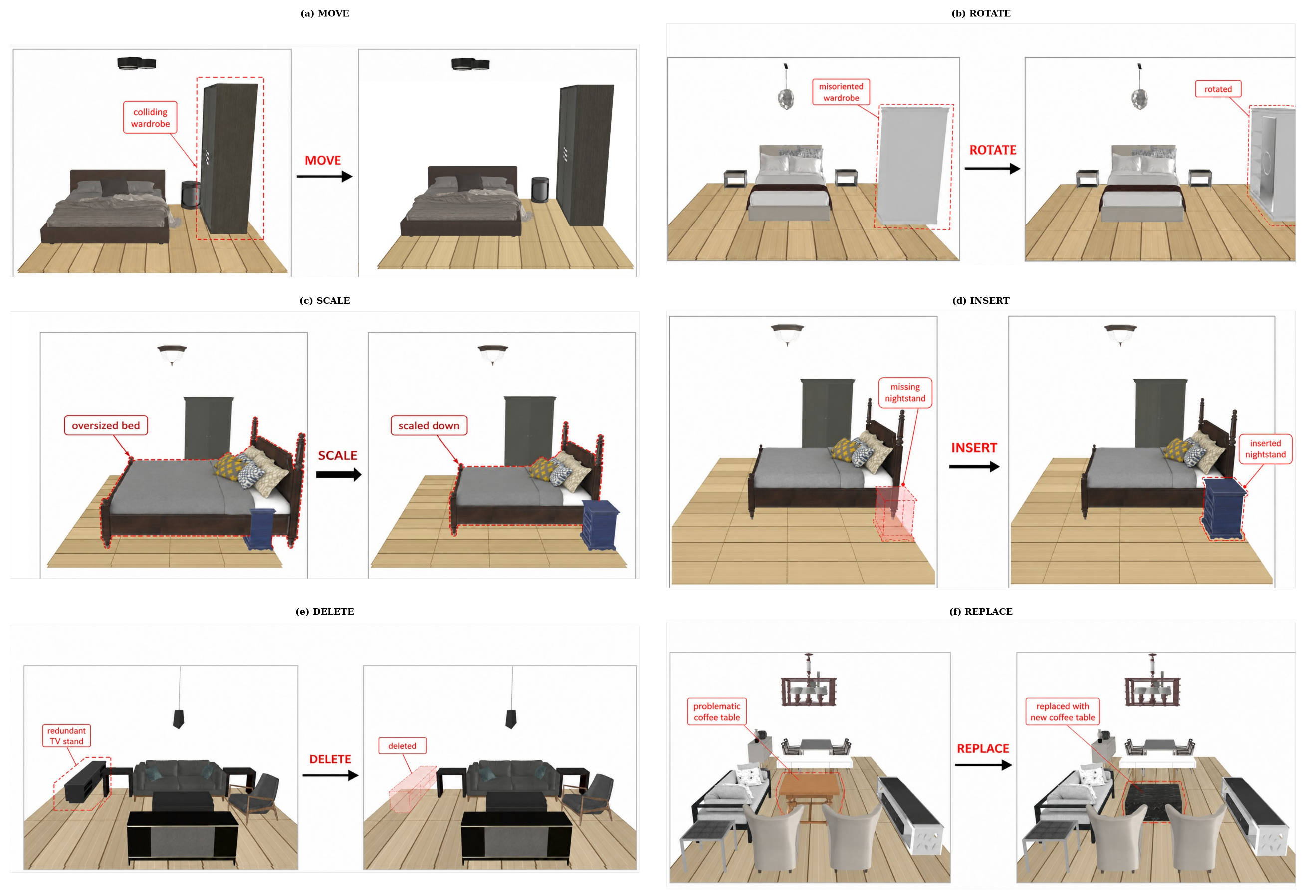}
    \caption{Representative before-and-after examples of the six StatePatch actions:
    \texttt{MOVE}, \texttt{ROTATE}, \texttt{SCALE}, \texttt{INSERT},
    \texttt{DELETE}, and \texttt{REPLACE}. Red annotations indicate the
    target object or affected region.}
    \label{fig:statepatch-action-examples}
\end{figure*}

\begin{figure*}[!t]
    \centering
    \includegraphics[width=\textwidth]{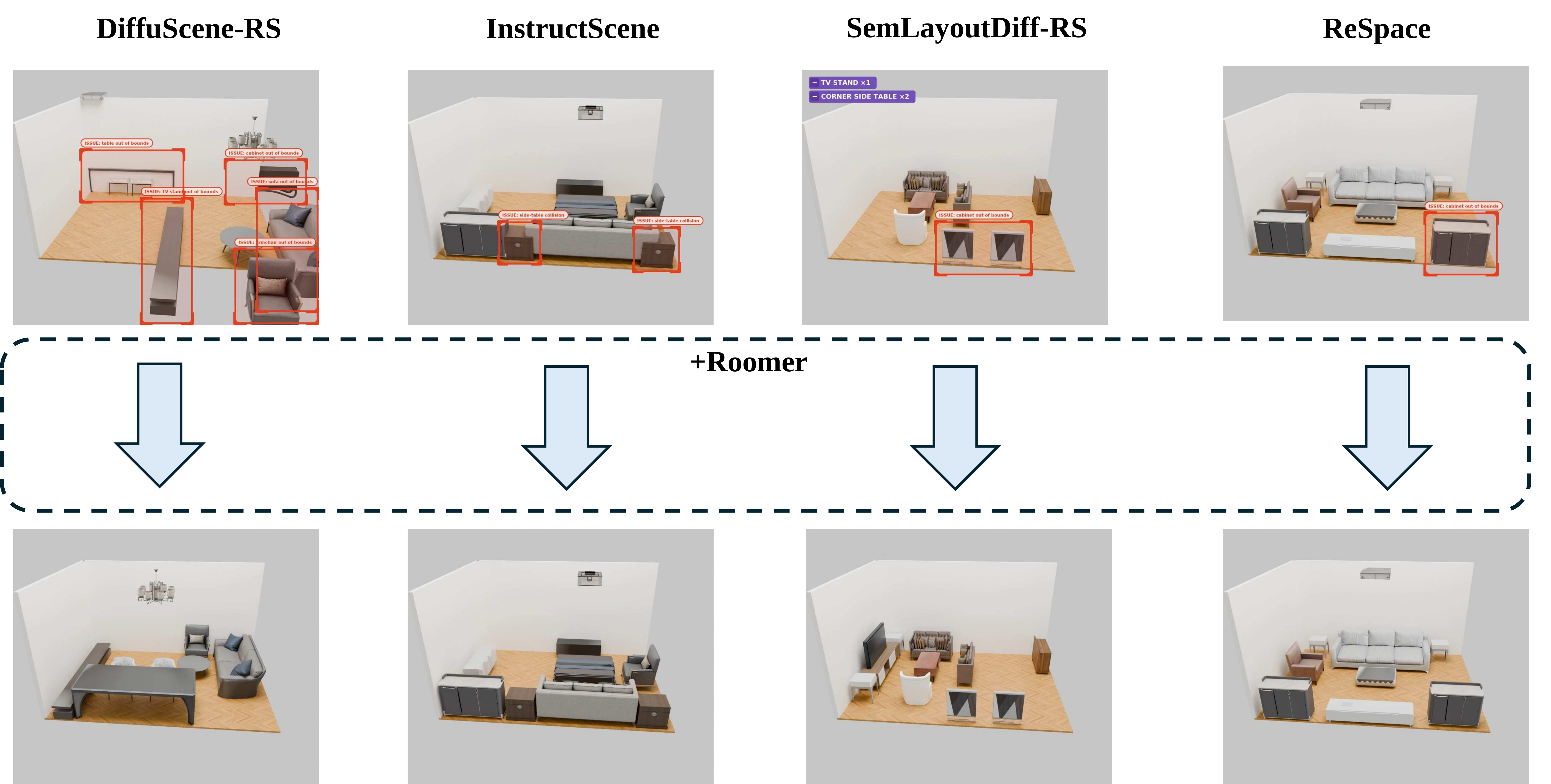}
    \caption{Plug-and-play refinement of outputs from four external generators.
    The top row shows frozen outputs from DiffuScene-RS, InstructScene,
    SemLayoutDiff-RS, and ReSpace with detected violations; the bottom row
    shows the corresponding layouts after applying Roomer. Roomer operates as
    a post-generation repair module without modifying the upstream generators.}
    \label{fig:pluggable-transfer}
\end{figure*}

\noindent\textbf{Supplementary organization.}
Figures~\ref{fig:statepatch-action-examples} and~\ref{fig:pluggable-transfer} provide an early visual overview of Roomer's action space and plug-and-play transfer. For more vivid demonstrations of the dynamic behavior of our method, please refer to the \textbf{supplementary videos} provided alongside this document. Appendices~A--H document the method and reproducibility chain, from instance-grounded context and Roomer-CC supervision to geometry conditioning, verification, deterministic search, and StatePatch learning; Appendices~I--M contain the main empirical support, including the complete evaluation protocol, Practical rules, controlled analyses, additional ablations, and professional validation.

\section{Instance-Grounded Repair Context}
\label{app:state-review}

\paragraph{Canonical RoState representation.}
At planning attempt $t$, Roomer encodes the current scene as the canonical RoState $\mathcal{S}^{t}=(\mathcal{A},\mathcal{Y}^{t},\mathcal{F}^{t},\Pi^{t})$, where $\mathcal{A}$ is immutable architecture, $\mathcal{Y}^{t}$ is the committed furniture layout, $\mathcal{F}^{t}$ contains geometry-derived functional regions, and $\Pi^{t}$ maps stable object references to concrete scene entities. These references preserve object grounding across the semantic rendering, numerical state, structured geometric evidence, planner output, and StatePatch execution, preventing a predicted edit from being applied to a different object instance.

\paragraph{Functional regions and temporary state updates.}
The current implementation constructs door-protection regions, bedside-clearance regions, dining-table operation regions, and primary circulation areas. These regions are derived from the current object geometry rather than stored as immutable annotations. Whenever an associated object is translated, rotated, or scaled, Roomer applies the candidate only to a temporary layout and reconstructs the affected functional regions before re-running the rule evaluator. INSERT, DELETE, and REPLACE similarly update the temporary entity registry and every functional region whose applicability or participating references change.

\paragraph{Instance-grounded RoReview construction.}
The evaluator converts every detected rule instance $v^{t}=(r^{t},\rho^{t})$ into the instance-grounded RoReview entry defined in the main paper. Its stable key $\kappa(v^{t})$ links the serialized entry to the underlying rule instance; $\tau^{t}$ records the violation type, $\mathcal{E}^{t}$ contains implicated furniture, architectural elements, and constructed functional regions, $\mathcal{W}^{t}$ assigns roles such as functional anchor, protected region, obstructing object, or expected category, and $\boldsymbol{\delta}^{t}$ records continuous geometric evidence. Depending on the rule, this evidence includes clearance deficit, intrusion or overlap ratio, overlap depth or area, out-of-bounds distance or area, opening-region intrusion, angular deficit, or passage-width deficit.

For a bedside-clearance violation, the bed is the functional anchor, the generated side-clearance region is protected, and furniture intersecting that region is identified as an obstructing object. For a door-blocking violation, the architectural opening and its swing proxy are protected, while intersecting furniture is implicated as the obstruction. RoReview therefore binds each measured violation to the concrete entities and relational roles needed for local repair, but deliberately does not prescribe the StatePatch target or action; that semantic decision remains the planner's responsibility. Stable issue identities, severity normalization, scheduling, and blocking behavior are specified in Appendices~\ref{app:verification} and~\ref{app:solver-control}.

\section{Roomer-CC: Controlled-Corruption Supervision and Data Provenance}
\label{app:data-partitions}

\paragraph{Controlled-corruption generation.}
Roomer-CC starts from valid 3D-FRONT layouts that satisfy their target room specifications. Each source layout generates multiple independent outer corruption proposals. For every proposal, the pipeline selects one operator from \texttt{MOVE}, \texttt{ROTATE}, \texttt{SCALE}, \texttt{INSERT}, \texttt{DELETE}, or \texttt{REPLACE} and modifies position, orientation, footprint scale, object presence, or object category. The corrupted layout is rendered and converted into a fresh RoState, RoReview, and structured evidence set; no evidence is copied from the valid source state. Construction-time verification determines which proposals become retained Roomer-CC instances, yielding the split-specific multiplicities reported in Table~\ref{tab:roomer-cc-splits}.

\paragraph{Verified inverse StatePatch supervision.}
Each retained instance pairs the corrupted observation and RoState with instance-grounded violation evidence and a known-feasible inverse StatePatch $a=c^{-1}$. INSERT and DELETE corruptions are supervised by DELETE and INSERT StatePatches, respectively, while MOVE, ROTATE, SCALE, and REPLACE use inverse parameters or the original source attributes. For a DELETE corruption, the INSERT target is represented through action-specific \texttt{new\_instance} fields rather than a reference to an observed object. Construction-time verification retains an instance only when (i) the intended violation is triggered, (ii) the implicated object or missing role is correctly identified, and (iii) the inverse StatePatch resolves the target violation without introducing a new hard violation or breaking a protected satisfied relation. The executable target is serialized under the same StatePatch schema used during inference.

\paragraph{Construction attempts and retention.}
A construction \emph{attempt} is one outer corruption proposal. For examples whose inverse repair action is \texttt{DELETE}, construction applies an \texttt{INSERT} corruption and evaluates up to 120 internal placement samples for the inserted object within the same outer proposal. These internal samples therefore count as one attempt rather than 120 attempts. Table~\ref{tab:roomer-cc-construction} reports the construction flow for the 61,010-instance Roomer-CC training split. \emph{Retained/Attempts} is the end-to-end retention rate.

\begin{table*}[!t]
\centering
\small
\setlength{\tabcolsep}{4.0pt}
\begin{tabular*}{\textwidth}{@{\extracolsep{\fill}}lccccc@{}}
\toprule
\textbf{Repair action} & \textbf{Attempts} & \textbf{Triggered} & \textbf{Correctly attributed} & \textbf{Retained} & \textbf{Retained/Attempts} \\
\midrule
\texttt{INSERT}  & 4,295  & 4,188  & 4,167  & 3,951  & 92.0\% \\
\texttt{DELETE}  & 18,543 & 17,060 & 16,719 & 13,722 & 74.0\% \\
\texttt{MOVE}    & 25,103 & 19,580 & 18,797 & 16,317 & 65.0\% \\
\texttt{REPLACE} & 15,245 & 13,873 & 13,734 & 12,958 & 85.0\% \\
\texttt{ROTATE}  & 16,675 & 14,674 & 14,381 & 13,340 & 80.0\% \\
\texttt{SCALE}   & 2,006  & 1,043  & 991    & 722    & 36.0\% \\
\midrule
\textbf{Total} & \textbf{81,867} & \textbf{70,418} & \textbf{68,789} & \textbf{61,010} & \textbf{74.5\%} \\
\bottomrule
\end{tabular*}
\caption{Roomer-CC construction statistics for the training split. Rows are indexed by the inverse repair action; for example, \texttt{DELETE} supervision is generated by \texttt{INSERT} corruptions. An attempt is one outer corruption proposal, and internal placement sampling does not increase the attempt count.}
\label{tab:roomer-cc-construction}
\end{table*}

The corresponding stage-wise rates are
\begin{equation}
\begin{aligned}
\frac{70{,}418}{81{,}867} &= 86.02\%,\\
\frac{68{,}789}{70{,}418} &= 97.69\%,\\
\frac{61{,}010}{68{,}789} &= 88.69\%.
\end{aligned}
\label{eq:roomer-cc-stage-retention}
\end{equation}
The end-to-end retention rate is
\begin{equation}
\operatorname{Retention}
=
\frac{\text{Retained}}{\text{Attempts}}.
\label{eq:roomer-cc-end-to-end-retention}
\end{equation}

\paragraph{Severity distribution of retained corruptions.}
Table~\ref{tab:roomer-cc-severity} summarizes the normalized severity of retained Roomer-CC training corruptions. \emph{Door-access violation} denotes the training-time door-zone usability failure measured by actual blocking severity; the final Roomer-Eval metric instead uses the frozen door swing proxy defined in Appendix~\ref{app:roomer-eval-rules}. The wrong-orientation distribution is right-skewed: its mean exceeds its upper quartile because of a long high-severity tail.

\begin{table*}[!t]
\centering
\small
\setlength{\tabcolsep}{7.0pt}
\begin{tabularx}{\textwidth}{@{}>{\raggedright\arraybackslash}p{0.32\textwidth}*{4}{>{\centering\arraybackslash}X}@{}}
\toprule
\textbf{Issue family} & \textbf{Mean} & \textbf{Median} & \textbf{Q25--Q75} & \textbf{P90} \\
\midrule
Collision & 0.19 & 0.14 & 0.07--0.25 & 0.43 \\
Out of bounds & 0.25 & 0.19 & 0.09--0.35 & 0.58 \\
Door-access violation & 0.29 & 0.22 & 0.10--0.40 & 0.64 \\
Window blocking & 0.24 & 0.18 & 0.08--0.34 & 0.57 \\
Wrong orientation & 0.14 & 0.06 & 0.03--0.10 & 0.49 \\
Bedside clearance (Roomer-CC) & 0.35 & 0.29 & 0.15--0.49 & 0.72 \\
Dining clearance & 0.31 & 0.26 & 0.13--0.46 & 0.67 \\
Living functional angle & 0.24 & 0.19 & 0.09--0.34 & 0.52 \\
Walkable disconnected & 0.40 & 0.35 & 0.18--0.59 & 0.80 \\
Relation distance & 0.29 & 0.23 & 0.11--0.42 & 0.66 \\
\bottomrule
\end{tabularx}
\caption{Normalized severity distribution of retained Roomer-CC corruptions in the training split.}
\label{tab:roomer-cc-severity}
\end{table*}

\paragraph{Binary content-rule severity.}
Binary content failures use unit severity. Before RoReview construction, every object-category or object-type mismatch is canonicalized as \texttt{wrong\_object\_category\_type}. A single object instance therefore contributes at most one category/type issue key.
\begin{table}[!t]
\centering
\small
\setlength{\tabcolsep}{4.0pt}
\begin{tabularx}{\columnwidth}{@{}>{\raggedright\arraybackslash}X>{\centering\arraybackslash}p{0.22\columnwidth}@{}}
\toprule
\textbf{Issue} & \textbf{Severity} \\
\midrule
Missing required object & 1.0 \\
Extra object & 1.0 \\
Wrong object category/type & 1.0 \\
\bottomrule
\end{tabularx}
\caption{Severity assigned to binary Roomer-CC content rules.}
\label{tab:binary-content-severity}
\end{table}

\paragraph{Roomer-CC and Roomer-Eval bedside definitions.}
Roomer-CC corruption construction uses the fixed bedside-clearance threshold
\begin{equation}
c_{\mathrm{usable}}\geq0.45\,\mathrm{m},
\end{equation}
with severity
\begin{equation}
s_{\mathrm{CC}}
=
\operatorname{clip}\!\left(
\frac{0.45-c_{\mathrm{usable}}}{0.45},0,1
\right).
\label{eq:roomer-cc-bedside-severity}
\end{equation}
Roomer-Eval instead scales the requirement by bed width:
\begin{equation}
\frac{c_{\mathrm{usable}}}{w_{\mathrm{bed}}}\geq0.30,
\end{equation}
with severity
\begin{equation}
s_{\mathrm{Eval}}
=
\operatorname{clip}\!\left(
\frac{0.30-c_{\mathrm{usable}}/w_{\mathrm{bed}}}{0.30},0,1
\right).
\label{eq:roomer-eval-bedside-severity}
\end{equation}
The two severities are computed and reported separately. The final Practical metric consists only of bedside ratio, dining clearance, living functional angle, door proxy, and walkable connectivity. Collision, out-of-bounds placement, window blocking, and binary content rules are reported separately and do not enter the five Practical families.

\paragraph{Room-level data partitions.}

The data pipeline uses hierarchical room-level partitions. The Qwen-Image train, validation, and test splits are disjoint. Roomer-CC is constructed exclusively from the 3,952 Qwen-Image training rooms, which are further partitioned before corruption generation so that a source room and all derived corruptions remain in one planner split. Its test split is therefore held out from vision-language model planner training and configuration selection, whereas the separate Qwen-Image test split forms \textit{common-1100} and is excluded from all training and tuning.

\begin{table}[!t]
\centering
\small
\setlength{\tabcolsep}{2.5pt}
\begin{tabular*}{\columnwidth}{@{\extracolsep{\fill}}lcccc@{}}
\toprule
\textbf{Split} & \textbf{Bedroom} & \textbf{Living room} & \textbf{Dining room} & \textbf{Total} \\
\midrule
Train & 2,822 & 543 & 587 & 3,952 \\
Validation & 409 & 77 & 85 & 571 \\
Test & 777 & 155 & 168 & 1,100 \\
\bottomrule
\end{tabular*}
\caption{Qwen-Image room-level partitions.}
\label{tab:qwen-image-splits}
\end{table}

The Qwen-Image training split is used for model training and as the Roomer-CC source-room pool. Its validation split is used for Qwen-Image checkpoint selection and system hyperparameter development. The test split is the frozen \textit{common-1100} cohort and is not used for training or tuning.

\begin{table*}[!t]
\centering
\small
\setlength{\tabcolsep}{5.0pt}
\begin{tabularx}{\textwidth}{@{}>{\raggedright\arraybackslash}p{0.14\textwidth}*{6}{>{\centering\arraybackslash}X}@{}}
\toprule
\textbf{Split} & \textbf{Bedroom} & \textbf{Living room} & \textbf{Dining room} & \textbf{Source rooms} & \textbf{Instances} & \textbf{Inst./room} \\
\midrule
Train & 2,539 & 489 & 528 & 3,556 & 61,010 & 17.16 \\
Validation & 142 & 27 & 29 & 198 & 3,270 & 16.52 \\
Test & 141 & 27 & 30 & 198 & 3,270 & 16.52 \\
\midrule
Total & 2,822 & 543 & 587 & 3,952 & 67,550 & 17.09 \\
\bottomrule
\end{tabularx}
\caption{Roomer-CC source-room partitions and verified corruption instances.}
\label{tab:roomer-cc-splits}
\end{table*}

The Roomer-CC train/validation/test source-room proportions are 89.98\%/5.01\%/5.01\%. The training split is used for vision-language model planner parameter learning, the validation split selects checkpoints and planner configurations, and the test split remains fully held out from training and configuration selection. Corruption multiplicity varies slightly because construction-time verification removes instances that fail to trigger the intended violation or whose paired repair fails verification.

\section{Qwen-Image Training and Optimization Details}
\label{app:qwen-image-training}

Qwen-Image is fine-tuned to map textual furniture requirements and architectural condition images to $512\times512$ semantic furniture layouts. The fine-tuning corpus is exactly the 3,952-room Qwen-Image training split in Table~\ref{tab:qwen-image-splits}, comprising 2,822 bedrooms, 543 living rooms, and 587 dining rooms. The 571-room validation split is used only for checkpoint selection, and no room from the frozen \textit{common-1100} test split contributes a fine-tuning example. Only LoRA adapters attached to the diffusion transformer (DiT) are trainable; the base-model weights remain frozen.

\begin{table*}[!t]
\centering
\small
\setlength{\tabcolsep}{4.0pt}
\begin{tabularx}{\textwidth}{@{}>{\raggedright\arraybackslash}p{0.19\textwidth}>{\raggedright\arraybackslash}p{0.29\textwidth}>{\raggedright\arraybackslash}p{0.19\textwidth}>{\raggedright\arraybackslash}X@{}}
\toprule
\textbf{Configuration} & \textbf{Value} & \textbf{Configuration} & \textbf{Value} \\
\midrule
Base model & Qwen-Image & Training layouts & 3,952 \\
Room-type distribution & 2,822 bedroom / 543 living room / 587 dining room & Input & Textual furniture requirements + architectural condition image \\
Output & $512\times512$ semantic furniture layout & Trainable modules & DiT LoRA only \\
LoRA rank / alpha / dropout & $64$ / $64$ / $0$ & Trainable parameters & 471,859,200 \\
Optimizer & AdamW & Weight decay & $0.01$ \\
Batch size / accumulation & $1$ / $1$ & Gradient clipping / warmup & none / none \\
Precision & BF16 computation; frozen base weights in FP8 & Hardware & $1\times$ NVIDIA A800 80GB \\
Stage 1 schedule & Constant learning rate of $5\times10^{-5}$ for 1 epoch & Stage 1 updates & 3,952 \\
Stage 2 schedule & Cosine decay from $2\times10^{-5}$ to $2\times10^{-6}$ for 3 epochs & Stage 2 updates & 11,856 \\
Total training & 4 epochs / 15,808 optimizer updates & Checkpoint selection & best validation performance on the 571-room validation split \\
\bottomrule
\end{tabularx}
\caption{Complete Qwen-Image fine-tuning configuration. The two stages are executed sequentially: constant-rate adaptation followed by lower-rate cosine refinement.}
\label{tab:qwen-image-training}
\end{table*}

\subsection{Qwen Parsing and 3D Assembly}
\label{app:qwen-parsing-assembly}

The Qwen-Image output is decoded into structured furniture instances and assembled with category-compatible 3D assets. The attribute-prediction module (APM) directly resolves yaw for orientation-sensitive objects whenever possible; unresolved cases invoke the deterministic orientation fallback. Table~\ref{tab:qwen-parsing-assembly} reports the complete parsing, retrieval, and assembly statistics on \textit{common-1100}. Statistical units are stated explicitly because scene-, component-, object-, and query-level rates are not interchangeable.

\begin{table*}[!t]
\centering
\small
\setlength{\tabcolsep}{6.0pt}
\begin{tabularx}{\textwidth}{@{}>{\raggedright\arraybackslash}X>{\centering\arraybackslash}p{0.18\textwidth}>{\raggedright\arraybackslash}p{0.28\textwidth}@{}}
\toprule
\textbf{Stage} & \textbf{Value} & \textbf{Statistical unit} \\
\midrule
Palette decode rate & 100.00\% & Scene level \\
Structural parse rate & 99.82\% & Scene level: 1,098/1,100 \\
Component retention rate & 98.60\% & Component level \\
APM direct-yaw resolution rate & 98.40\% & Orientation-required objects \\
Orientation fallback invocation rate & 1.60\% & Orientation-required objects \\
Same-category asset retrieval success & 99.20\% & Asset queries \\
Distortion-blocked object rate & 0.55\% & Asset queries \\
Empty-category candidate rate & 0.20\% & Asset queries \\
Other retrieval failure rate & 0.05\% & Asset queries \\
\midrule
\textbf{Raw scene assembly valid} & \textbf{99.09\%} & \textbf{1,090/1,100 scenes} \\
\textbf{Primary frozen-output valid} & \textbf{99.82\%} & \textbf{1,098/1,100 scenes} \\
\textbf{Final assembly valid after fallback} & \textbf{100.00\%} & \textbf{1,100/1,100 scenes} \\
\bottomrule
\end{tabularx}
\caption{Qwen parsing, same-category asset retrieval, and 3D assembly statistics on \textit{common-1100}.}
\label{tab:qwen-parsing-assembly}
\end{table*}

The raw scene-assembly validity is
\begin{equation}
\frac{1{,}090}{1{,}100}=99.0909\%\approx99.09\%.
\end{equation}
The deterministic recovery path is
\begin{equation}
1{,}090
\xrightarrow{\substack{\text{object-level}\\\text{fallback}}}
1{,}098
\xrightarrow{\substack{\text{scene-level frozen-candidate}\\\text{fallback}}}
1{,}100.
\label{eq:assembly-recovery-path}
\end{equation}
Thus, raw assembly directly succeeds for 1,090 scenes, deterministic object-level fallback recovers eight scenes, and traversal of the frozen scene-level candidate pool recovers the remaining two scenes.

The asset-query outcomes are mutually exclusive and exhaustive:
\begin{equation}
\begin{aligned}
100\%-99.20\% &= 0.80\%,\\
0.80\% &= 0.55\%+0.20\%+0.05\%.
\end{aligned}
\end{equation}
or equivalently
\begin{equation}
99.20\%+0.55\%+0.20\%+0.05\%=100.00\%.
\end{equation}

\section{Geometry-Conditioning Implementation}
\label{app:geometry-conditioning}

\paragraph{Aligned semantic and geometric repair context.}
For each active violation, the semantic stream contains the current top-down semantic rendering $I^{t}$ of $\mathcal{Y}^{t}$, the target room specification $\mathcal{G}$ with its textual user requirements, the complete human-readable RoState, and the selected instance-grounded RoReview entry. Structured geometric evidence is encoded separately but uses the same stable object references as the textual state and StatePatch schema. Consequently, object grounding remains aligned across rendered regions, human-readable descriptions, evidence tokens, and executable edit targets.

\paragraph{Evidence entities.}
Each RoReview retains at most $N_e=16$ evidence entities. If more are available, we retain (1) functional anchors, (2) obstructing objects, (3) protected architectural elements or functional regions, and (4) neighboring objects ordered by violation relevance and geometric distance. Remaining positions are padded and masked in every attention operation.

For room width $W$, depth $D$, and lower floor-plane bounds $(x_{\min},z_{\min})$, numerical geometry is normalized as
\begin{equation}
\widetilde{x}=\frac{x-x_{\min}}{W},
\qquad
\widetilde{z}=\frac{z-z_{\min}}{D}.
\end{equation}
\begin{equation}
\widetilde{w}=\frac{w}{W},
\qquad
\widetilde{d}=\frac{d}{D}.
\end{equation}
\begin{equation}
\widetilde{d}_{ij}=\frac{d_{ij}}{\sqrt{W^2+D^2}}.
\end{equation}
Yaw is represented by $(\sin\theta,\cos\theta)$ to avoid the discontinuity between $-180^{\circ}$ and $180^{\circ}$. Clearance deficit, intrusion ratio, and overlap ratio use their rule-specific values already normalized to $[0,1]$. Each entity token concatenates learned embeddings for category, entity type, relational role, violation type, and editability with normalized center, footprint size, yaw sine/cosine, pairwise distance, and type-specific violation measurements; the result is projected to $d_e=512$ and normalized before aggregation.

\begin{table}[!t]
\centering
\small
\setlength{\tabcolsep}{4.0pt}
\begin{tabularx}{\columnwidth}{@{}>{\raggedright\arraybackslash}X>{\centering\arraybackslash}p{0.30\columnwidth}@{}}
\toprule
\textbf{Component} & \textbf{Configuration} \\
\midrule
Maximum evidence entities & 16 \\
Entity-token dimension & 512 \\
Repair Queries $K_c$ & 8 \\
Aggregator layers & 2 \\
Attention heads & 8 \\
Per-head dimension & 64 \\
FFN dimension & 2048 \\
Attention dropout & 0 \\
FFN dropout & 0 \\
\bottomrule
\end{tabularx}
\caption{Geometry-conditioning configuration.}
\label{tab:geometry-conditioning}
\end{table}

\begin{table}[t]
\centering
\small
\setlength{\tabcolsep}{1mm}
\begin{tabular*}{\columnwidth}{@{\extracolsep{\fill}}lcrrr@{}}
\toprule
\textbf{Method} & \textbf{State} & \textbf{FID}$\downarrow$ & \textbf{KID} $\times 10^3\!\downarrow$ & \textbf{SCA Gap}$\downarrow$ \\
\midrule
\multirow{2}{*}{ReSpace}
& Before & 57.42 & \textbf{5.10} & 21.73 \\
& \cellcolor{ResultHighlight}\textbf{After}
& \cellcolor{ResultHighlight}\textbf{56.99}
& \cellcolor{ResultHighlight}6.13
& \cellcolor{ResultHighlight}\textbf{18.49} \\
\addlinespace[1pt]
\multirow{2}{*}{DiffuScene-RS}
& Before & 68.15 & 11.81 & \textbf{16.57} \\
& \cellcolor{ResultHighlight}\textbf{After}
& \cellcolor{ResultHighlight}\textbf{63.05}
& \cellcolor{ResultHighlight}\textbf{8.48}
& \cellcolor{ResultHighlight}22.08 \\
\addlinespace[1pt]
\multirow{2}{*}{InstructScene}
& Before & 64.90 & 7.21 & 19.98 \\
& \cellcolor{ResultHighlight}\textbf{After}
& \cellcolor{ResultHighlight}\textbf{56.06}
& \cellcolor{ResultHighlight}\textbf{2.64}
& \cellcolor{ResultHighlight}\textbf{17.88} \\
\addlinespace[1pt]
\multirow{2}{*}{SemLayoutDiff-RS}
& Before & 90.68 & 29.68 & 34.12 \\
& \cellcolor{ResultHighlight}\textbf{After}
& \cellcolor{ResultHighlight}\textbf{80.57}
& \cellcolor{ResultHighlight}\textbf{18.75}
& \cellcolor{ResultHighlight}\textbf{26.49} \\
\bottomrule
\end{tabular*}
\caption{Paired distributional-quality results before and after applying Roomer to external generators on \textit{common-1100}. Ten of the twelve paired comparisons improve. Light shading denotes the refined state, bold indicates the better value within each Before--After pair, KID is reported as $\times 10^3$, and all 1,100 scenes are retained.}
\label{tab:generator-transfer-distribution}
\end{table}

\paragraph{Decoder injection.}
Qwen2.5-VL-7B-Instruct has 28 language-decoder layers with hidden size 3,584. Geometry-conditioning adapters are inserted at 1-based layers $\{22,24,26,28\}$, corresponding to zero-based implementation indices $[21,23,25,27]$. Each geometry-conditioning adapter uses eight-head, 512-dimensional cross-attention. Language states are projected from 3,584 to 512 dimensions before cross-attention and projected back to 3,584 dimensions before residual addition.

\paragraph{Initialization.}
Learned evidence embeddings and Repair Queries are sampled from $\mathcal{N}(0,0.02^2)$. Numerical and evidence-output projections, aggregator attention and FFN matrices, and each geometry-conditioning adapter's query and cross-attention matrices use Xavier-uniform initialization; all associated biases are zero, and LayerNorm weights and biases are initialized to one and zero. For every injected layer, $\mathbf{W}_{\ell}^{o}=\mathbf{0}$ and $\gamma_{\ell}=1$. Thus, the geometry branch initially contributes zero and the modified network matches the pretrained backbone, while the nonzero gate preserves gradient flow to the output projection.

\section{Training and Optimization Details}
\label{app:training-details}

The base language and vision weights remain frozen throughout training. AdamW uses four parameter groups: LoRA parameters use learning rate $10^{-5}$ and zero weight decay; geometry matrix weights use learning rate $10^{-4}$ and weight decay $0.01$; geometry biases, normalization parameters, and Repair Queries use learning rate $10^{-4}$ and zero weight decay; residual gates use the same learning rate and zero weight decay. Geometry matrix weights include the evidence-output projection, aggregator attention and FFN projections, and the geometry-conditioning adapter query, cross-attention, and output matrices.

\begin{table*}[!t]
\centering
\small
\setlength{\tabcolsep}{4.0pt}
\begin{tabularx}{\textwidth}{@{}>{\raggedright\arraybackslash}p{0.19\textwidth}>{\raggedright\arraybackslash}p{0.29\textwidth}>{\raggedright\arraybackslash}p{0.19\textwidth}>{\raggedright\arraybackslash}X@{}}
\toprule
\textbf{Configuration} & \textbf{Value} & \textbf{Configuration} & \textbf{Value} \\
\midrule
Base model & Qwen2.5-VL-7B-Instruct & Roomer-CC instances & 61,010 / 3,270 / 3,270 \\
LoRA rank / alpha / dropout & $8$ / $16$ / $0.05$ & LoRA targets & \texttt{q\_proj}, \texttt{k\_proj}, \texttt{v\_proj}, \texttt{o\_proj} \\
Trainable LoRA parameters & 5.046M & Trainable geometry parameters & approximately 27.446M \\
Total trainable parameters & approximately 32.492M & Frozen modules & vision encoder, visual merger, original LM weights \\
Trainable geometry modules & evidence encoder, Repair Queries, query aggregator, geometry-conditioning adapters & Optimizer & AdamW (\texttt{adamw\_torch}) \\
LoRA learning rate / decay & $10^{-5}$ / $0$ & Geometry learning rate & $10^{-4}$ \\
Geometry matrix decay & $0.01$ & Bias/norm/query/gate decay & $0$ \\
AdamW betas / epsilon & $(0.9,0.999)$ / $10^{-8}$ & Gradient clipping & $1.0$ \\
Scheduler / warmup & cosine / 5\% (477 steps) & Epochs / optimizer steps & 5 / 9,535 \\
Per-GPU batch / GPUs & 4 / 2 & Accumulation / effective batch & 4 / 32 \\
Maximum sequence length & 8,192 & Precision & BF16; FP16 disabled \\
Gradient checkpointing & enabled & DeepSpeed & ZeRO-3; no offload \\
Hardware & $2\times$ NVIDIA A800-SXM4-80GB & Random seed & 42 \\
Checkpoint selection & best validation repair performance & Selected checkpoint & step 9,535 \\
\bottomrule
\end{tabularx}
\caption{Complete vision-language model planner training configuration. Source rooms are partitioned before corruption generation, preventing a room and its derived corruptions from crossing splits.}
\label{tab:training-configuration}
\end{table*}

\section{Verification-Gated Commit: Formal Definitions}
\label{app:verification}

\paragraph{Rule instances and normalized severity.}
Let $\mathcal{R}$ contain all active repair-rule families, including hard, content, relational, and practical violations. For each rule family $r\in\mathcal{R}$, let $\mathcal{I}_{r}(\mathcal{Y};\mathcal{A},\mathcal{G})$ be its applicable instance set. Each instance is $v=(r,\rho)$, where $\rho$ contains stable entity references and $\operatorname{refs}(v)$ denotes the participating reference set. Its normalized severity satisfies $\bar e_v(\mathcal{Y};\mathcal{A},\mathcal{G})\in[0,1]$, with zero denoting satisfaction. Representative continuous definitions are
\begin{equation}
\bar e_v^{\mathrm{clear}}(\mathcal{Y};\mathcal{A},\mathcal{G})
=
\operatorname{clip}\!\left(
\frac{c_{\mathrm{req}}-c_{\mathrm{obs}}}{c_{\mathrm{req}}},0,1
\right).
\end{equation}
\begin{equation}
\bar e_v^{\mathrm{OOB}}(\mathcal{Y};\mathcal{A},\mathcal{G})
=
\operatorname{clip}\!\left(
\frac{A_{\mathrm{outside}}}{A_{\mathrm{footprint}}},0,1
\right).
\end{equation}
\begin{equation}
\bar e_v^{\mathrm{angle}}(\mathcal{Y};\mathcal{A},\mathcal{G})
=
\operatorname{clip}\!\left(
\frac{\theta_{\mathrm{req}}-\theta_{\mathrm{obs}}}{\theta_{\mathrm{req}}},0,1
\right).
\end{equation}
Binary composition or category violations use $\bar e_v\in\{0,1\}$. The stable issue key is $\kappa(v)=(r,\operatorname{sort}(\rho))$; the implementation serializes it as \texttt{issue\_type|sorted\_refs}.

\paragraph{Stable hard-violation keys.}
Let $\mathcal{R}_{\mathrm{hard}}$ contain collision, out-of-floor-boundary, and architectural-opening blockage families. The hard-violation key set is
\begin{equation}
\mathcal{H}(\mathcal{Y};\mathcal{A},\mathcal{G})
=
\left\{
\kappa(v)\;\middle|\;
\begin{gathered}
r\in\mathcal{R}_{\mathrm{hard}},\\
v\in\mathcal{I}_{r}(\mathcal{Y};\mathcal{A},\mathcal{G}),\\
\bar e_v(\mathcal{Y};\mathcal{A},\mathcal{G})>0
\end{gathered}
\right\}.
\label{eq:hard-key-set-app}
\end{equation}
Using stable keys rather than only a hard-error count ensures that a candidate cannot remove one hard violation while introducing a different one.

\paragraph{Verifier-derived protected satisfied relations.}
Protection metadata is derived deterministically by the verifier and is not predicted by the planner. Let $\mathcal{O}^{t}$ be the set of stable references of mutable furniture entities in the committed layout. For a candidate transaction $p$, define
\begin{equation}
\begin{aligned}
T_t(p)
&=\operatorname{ChangedRefs}(p)\cap\mathcal{O}^{t},\\
P_t(p)
&=\mathcal{O}^{t}\setminus T_t(p).
\end{aligned}
\label{eq:protected-reference-sets-app}
\end{equation}
where $T_t(p)$ contains every existing furniture reference whose state is modified or deleted. A standard \texttt{MOVE}, \texttt{ROTATE}, \texttt{SCALE}, \texttt{REPLACE}, or \texttt{DELETE} StatePatch therefore has one changed existing reference, whereas \texttt{INSERT} has $T_t(p)=\varnothing$ because it does not alter an existing entity. The action and target are fixed across the seed and its deterministic fallback neighborhood, so $T_t$ and $P_t$ are constant within a standard repair transaction and the candidate argument is omitted below.

For a rule instance $v$, let $\operatorname{deps}(v)\subseteq\mathcal{O}^{t}$ denote the mutable furniture references on which its geometry or semantics depend. This dependency set includes the anchor objects of any derived functional region, so a clearance or circulation relation is not treated as independent merely because the region itself has a synthetic reference. Define
\begin{equation}
\mathcal{Z}_{\mathrm{keep}}^{t}
=
\left\{
\kappa(v)\;\middle|\;
\begin{gathered}
v\in\displaystyle\bigcup_{r\in\mathcal{R}}
\mathcal{I}_{r}(\mathcal{Y}^{t};\mathcal{A},\mathcal{G}),\\
\bar e_v(\mathcal{Y}^{t};\mathcal{A},\mathcal{G})=0,\\
\operatorname{deps}(v)\subseteq P_t
\end{gathered}
\right\}.
\label{eq:protected-relations-app}
\end{equation}
and
\begin{equation}
\mathcal{Z}(\mathcal{Y};\mathcal{A},\mathcal{G})
=
\left\{
\kappa(v)\;\middle|\;
\begin{gathered}
v\in\displaystyle\bigcup_{r\in\mathcal{R}}
\mathcal{I}_{r}(\mathcal{Y};\mathcal{A},\mathcal{G}),\\
\bar e_v(\mathcal{Y};\mathcal{A},\mathcal{G})=0
\end{gathered}
\right\}.
\end{equation}
Thus, every currently satisfied relation whose mutable dependencies are disjoint from the edited target must remain satisfied after the candidate is applied. Relations that necessarily depend on a moved, scaled, rotated, deleted, or replaced target are not inserted into $\mathcal{Z}_{\mathrm{keep}}^{t}$; their regressions are still controlled by the aggregate-residual condition. Target room-composition constraints are represented as content-rule instances and remain subject to the active-target, protected-relation, and aggregate-residual conditions rather than being folded into $\mathcal{C}$. The StatePatch schema remains the three-part action--target--parameter-seed representation defined in the main paper and contains no planner-supplied \texttt{protected\_refs} field.

\paragraph{Structural validity and aggregate residual.}
The indicator $\mathcal{C}(\mathcal{Y};\mathcal{A},\mathcal{G})=1$ iff all object attributes are finite, sizes are positive, references are unique, asset identifiers are category-compatible, object schemas are valid under $\mathcal{G}$, and the architecture matches the fixed environment $\mathcal{A}$. Action legality is enforced separately during schema, reference, and action validation before candidate instantiation. Structural validity is distinct from physical all-pass: $\mathcal{C}(\mathcal{Y};\mathcal{A},\mathcal{G})=1$ does not imply $\mathcal{H}(\mathcal{Y};\mathcal{A},\mathcal{G})=\varnothing$. Accordingly, reducing the active violation is insufficient unless the candidate passes the complete five-clause full-scene verification gate in main-paper Eq.~(3); this gate does not require every hard-violation key already present before the current transaction to be absent.

To prevent a family with many instances from dominating the residual, each family is first averaged over its applicable instances and all families receive equal weight. In the following display, $\mathcal{I}_{r}$ abbreviates $\mathcal{I}_{r}(\mathcal{Y};\mathcal{A},\mathcal{G})$:
\begin{equation}
\mathcal{V}(\mathcal{Y};\mathcal{A},\mathcal{G})
=
\frac{1}{|\mathcal{R}|}
\sum_{r\in\mathcal{R}}
\frac{
\displaystyle\sum_{v\in\mathcal{I}_{r}}
\bar e_v(\mathcal{Y};\mathcal{A},\mathcal{G})
}{
\displaystyle\max\!\left(1,|\mathcal{I}_{r}|\right)
}.
\label{eq:aggregate-residual-app}
\end{equation}
A family with no applicable instance contributes zero. This yields $\mathcal{V}\in[0,1]$ under uniform family weights $\lambda_r=1/|\mathcal{R}|$.

\paragraph{Exact target resolution and global descent.}
Let the recomputed unsatisfied-issue key set be
\begin{equation}
\mathcal{U}(\mathcal{Y};\mathcal{A},\mathcal{G})
=
\left\{
\kappa(v)\;\middle|\;
\begin{gathered}
r\in\mathcal{R},\\
v\in\mathcal{I}_{r}(\mathcal{Y};\mathcal{A},\mathcal{G}),\\
\bar e_v(\mathcal{Y};\mathcal{A},\mathcal{G})>0
\end{gathered}
\right\}.
\label{eq:unsatisfied-key-set-app}
\end{equation}
For the active rule instance $v^t$ represented by $q^t$, the target predicate used in main-paper Eq.~(3) is
\begin{equation}
\mathsf{Target}_{t}(p)
=
\mathbf{1}\!\left[
\kappa(v^t)\notin
\mathcal{U}(\mathcal{Y}_{p}^{t};\mathcal{A},\mathcal{G})
\right].
\label{eq:exact-target-resolution-app}
\end{equation}
Consequently, both continuously measured and binary targets must be fully resolved under the frozen rule evaluator before a candidate can be committed; a partial severity reduction is insufficient. Each detector applies its rule-specific geometric tolerance before emitting an issue, so Eq.~\eqref{eq:exact-target-resolution-app} tests absence of the stable issue key after full recomputation rather than exact equality of unthresholded floating-point geometry. The aggregate residual must additionally satisfy the main-paper condition with $\varepsilon_V=10^{-6}$, which enforces strict global descent while avoiding rejection caused only by floating-point noise after family-wise averaging.

\paragraph{Correspondence to main-paper schematic labels.}
Main-paper Fig.~2 uses compact visual shorthand. Its label ``Target improved'' denotes the detector-level target-resolution test in Eq.~\eqref{eq:exact-target-resolution-app}; ``Pro. rules preserved'' abbreviates the protected satisfied relations $\mathcal{Z}_{\mathrm{keep}}^{t}$ rather than all Practical rule instances; and ``No new violation'' denotes the no-new-hard-key condition $\mathcal{H}_{p}^{t}\setminus\mathcal{H}^{t}=\varnothing$. The figure's ``Repair plan $P_j^t$'' and ``Condition Token $Z_j^t$'' correspond, respectively, to the schema-constrained StatePatch proposal $\widehat p^t$ and the repair-query-conditioned geometry evidence $Z^t$. These schematic labels do not replace any clause of main-paper Eq.~(3).

\section{Deterministic Candidate Instantiation and Repair Scheduling}
\label{app:solver-control}

\paragraph{Planner-seed-first candidate ordering.}
After schema, reference, and action validation, the deterministic solver parses the raw autoregressive output and instantiates it as the normalized executable seed candidate $\widehat p^{t}$. The raw planner serialization is never committed directly. The normalized executable candidate is evaluated first; if it is rejected, the solver constructs the deterministic action-specific neighborhood
\begin{equation}
\mathcal{N}^{t}
=
\mathcal{N}_{\widehat a^t}
\left(
\widehat{\boldsymbol{\xi}}^{t},
\widehat{\boldsymbol{\alpha}}_{0}^{t};
q^{t},\mathcal{S}^{t}
\right)
\end{equation}
and orders its instantiated candidates as
\begin{equation}
\mathbf{B}^{t}
=
\left(b_1^{t},\ldots,b_{M_t}^{t}\right)
=
\operatorname{Order}
\left[
\operatorname{Inst}_{\widehat a^t}(\mathcal{N}^{t})
\right].
\label{eq:ordered-fallback-sequence-app}
\end{equation}
If the seed is rejected and at least one fallback passes verification, let $j^\star=\min\{j:\mathsf{Acc}_{t}(b_j^{t})=1\}$. The selected candidate is
\begin{equation}
p^{t,\star}
=
\begin{cases}
\widehat p^{t},
& \mathsf{Acc}_{t}(\widehat p^{t})=1,\\[1mm]
b_{j^\star}^{t},
& j^\star\text{ exists},\\[1mm]
\varnothing,
& \text{otherwise}.
\end{cases}
\label{eq:seed-first-short-circuit-app}
\end{equation}
Thus, when the normalized executable candidate instantiated from the planner seed passes verification, its resulting StatePatch is used without a fallback search. Deterministic numerical search is invoked only after that candidate is rejected and terminates at the first verified fallback. Fallback order is primarily determined by action-specific parameter distance from the seed and then by edit magnitude; the concrete sequences are listed below.

\paragraph{Action execution.}
All controlled variants use the same primitive action vocabulary $\mathbb{A}=\{\texttt{MOVE},\texttt{ROTATE},\texttt{SCALE},\texttt{INSERT},\texttt{DELETE},\texttt{REPLACE}\}$. Standard Roomer emits one target-grounded StatePatch per round. The Full-Scene JSON Rewrite ablation may propose multiple object changes in one transaction, but every difference from the committed state must be decomposable into these same six primitives; unsupported attribute changes and architecture edits are rejected. The protocol does not freeze object identity, category, count, or size; \texttt{INSERT}, \texttt{DELETE}, \texttt{REPLACE}, and \texttt{SCALE} may therefore alter these attributes, while room architecture is immutable.

Table~\ref{tab:action-execution} records the exact StatePatch scope, seed-first fallback sequence, and principal validity checks. The ordered search always evaluates the normalized executable candidate instantiated from the planner-predicted StatePatch seed before the listed deterministic fallbacks.

\begin{table*}[!t]
\centering
\small
\setlength{\tabcolsep}{3.2pt}
\begin{tabularx}{\textwidth}{@{}>{\raggedright\arraybackslash}p{0.075\textwidth}>{\raggedright\arraybackslash}p{0.205\textwidth}>{\raggedright\arraybackslash}p{0.32\textwidth}>{\raggedright\arraybackslash}X@{}}
\toprule
\vheadl{\textbf{Action}}
& \vheadl{\textbf{StatePatch scope}}
& \vheadl{\textbf{Ordered fallback}\\\textbf{after seed rejection}}
& \vheadl{\textbf{Principal constraints and}\\\textbf{implementation notes}} \\
\midrule
\texttt{INSERT} & Add one object through \texttt{new\_instance}. & Search a 0.1-m grid within 2.0 m of the proposed center. If no local candidate passes verification, search room-wide free space on a 0.2-m grid, capped at 1,000 positions. & Category must be nonempty; center values must be finite with per-axis magnitude at most 50 m; the footprint must lie inside the floor and overlap existing furniture by at most $0.005\,\mathrm{m}^2$. The required-category count must increase, the target \texttt{missing\_required\_object} issue must disappear, and no new hard or protected-relation violation may be introduced. \\
\texttt{DELETE} & Remove exactly one object referenced by \texttt{target\_ref}; no state-update fields are allowed. & No numerical search; evaluate the direct deletion only. & The target reference must exist uniquely; an optional \texttt{expected\_category} must match. This action addresses \texttt{extra\_object}; the target issue must disappear and no new hard-violation key may be introduced. \\
\texttt{MOVE} & Modify only \texttt{center\_m}; execution is normalized internally as \texttt{TRANSLATE}. & Evaluate the proposed displacement, then distances 0.05, 0.10, 0.15, 0.20, 0.30, and 0.40 m, followed by 0.60, 0.80, 1.00, and 1.20 m, with orthogonal offsets $\{-0.10,0,+0.10\}$ m. For collision, boundary, and opening-blockage targets, a room-wide 0.2-m grid is evaluated after all local candidates fail verification. & Eight cardinal/diagonal directions are supported. Proposed displacement components are bounded by 5 m, although room-wide internal candidates may exceed that displacement. The patch must remove the active target issue under the recomputed evaluator, and no new hard-violation key may be introduced; collision repair may not move implicated objects out of bounds. \\
\texttt{REPLACE} & Modify the same object's \texttt{category}, \texttt{asset\_id}, and/or \texttt{model\_jid}. & No continuous search; directly evaluate the semantic/asset substitution. & The target must exist uniquely. This action addresses \texttt{wrong\_object\_category\_type}; center, orientation, and footprint are retained, the resulting category must match the expected category, the active target issue must disappear, and 3D assembly must retrieve a compatible asset. \\
\texttt{ROTATE} & Modify \texttt{orientation\_deg} or \texttt{yaw\_rad}. & Evaluate the signed proposed offset, then fixed offsets $-180^{\circ}$, $-90^{\circ}$, $-45^{\circ}$, $+45^{\circ}$, $+90^{\circ}$, and $+180^{\circ}$, followed by local offsets $\Delta\theta_0+\{-15^{\circ},-5^{\circ},+5^{\circ},+15^{\circ}\}$. & Used for \texttt{wrong\_orientation} and for collision repair when rotation reduces overlap. Positive angles are clockwise in the top-down frame; the canonical signed seed satisfies $|\Delta\theta_0|\leq180^{\circ}$. Orientation must change, the active target issue must disappear, and no new hard-violation key may be introduced. Because yaw is normalized modulo $360^{\circ}$, the $-180^{\circ}$ and $+180^{\circ}$ offsets are state-equivalent; retaining both entries preserves the configured fallback sequence but does not enlarge the reachable candidate set. The configured \texttt{rotate\_snap\_angles\_deg} is not used by the candidate generator. \\
\texttt{SCALE} & Modify the two-dimensional footprint \texttt{size\_m}. & Evaluate \texttt{scale\_xz}, then standard factors $\{0.70,0.80,0.85,0.90,0.95\}$. Collision/blocking fallbacks search $0.60$--$0.95$ and remain subject to category-specific minimum footprints. & Each factor must be positive and no proposed factor may exceed 20. All candidates enforce category-specific minimum sizes; size must change, the active target issue must be removed under the recomputed evaluator, and no new hard-violation key may be introduced. \\
\bottomrule
\end{tabularx}
\caption{StatePatch execution domains and deterministic fallback behavior.}
\label{tab:action-execution}
\end{table*}

\paragraph{Action--issue coverage in Roomer-CC.}
All retained examples use the canonical issue taxonomy applied by RoReview and the scheduler. The 61,010-instance training split contains \texttt{INSERT} supervision for \texttt{missing\_required\_object}; \texttt{DELETE} for \texttt{extra\_object}; \texttt{REPLACE} for \texttt{wrong\_object\_category\_type}; \texttt{MOVE} for \texttt{collision}, \texttt{out\_of\_floor\_bounds}, \texttt{door\_blocking}, \texttt{window\_blocking}, \texttt{walkable\_disconnected}, \texttt{bed\_side\_clearance\_insufficient}, \texttt{dining\_table\_clearance\_insufficient}, \texttt{living\_sofa\_coffee\_tv\_angle}, \texttt{relation\_distance}, and \texttt{clearance\_insufficient}; \texttt{ROTATE} for \texttt{wrong\_orientation} and \texttt{collision}; and \texttt{SCALE} for \texttt{collision}, \texttt{door\_blocking}, \texttt{window\_blocking}, and \texttt{bed\_side\_clearance\_insufficient}. The inference action mask uses the same action--issue pairs.

\paragraph{Issue scheduling.}
The scheduler sorts detected issues by ascending integer priority and, within an equal-priority class, by descending implementation-level scheduler score. This tie-breaking score is distinct from the normalized rule severity $\bar e_v$ used by the verification gate and aggregate residual in Appendix~\ref{app:verification}. It is the maximum available diagnostic value among \texttt{intersection\_area\_m2}, \texttt{outside\_area\_m2}, \texttt{blocked\_ratio}, \texttt{overlap\_ratio}, \texttt{blocked\_area\_m2}, and \texttt{outside\_ratio}; remaining ties preserve detector order. Because integer priority is applied first, this raw diagnostic score is not compared across different issue families and does not enter $\mathcal V$.

\begin{table}[!t]
\centering
\small
\setlength{\tabcolsep}{4.0pt}
\begin{tabularx}{\columnwidth}{@{}>{\raggedright\arraybackslash}X>{\centering\arraybackslash}p{0.18\columnwidth}@{}}
\toprule
\textbf{Issue} & \textbf{Priority} \\
\midrule
\texttt{extra\_object} & 9 \\
\texttt{collision} & 10 \\
\texttt{out\_of\_floor\_bounds} & 20 \\
\texttt{door\_blocking} & 30 \\
\texttt{window\_blocking} & 40 \\
\texttt{walkable\_disconnected} & 50 \\
\texttt{wrong\_orientation} & 100 \\
\texttt{bed\_side\_clearance\_insufficient} & 110 \\
\texttt{dining\_table\_clearance\_}\allowbreak\texttt{insufficient} & 115 \\
\texttt{living\_sofa\_coffee\_tv\_angle\ldots} & 120 \\
\texttt{relation\_distance} & 130 \\
\texttt{clearance\_insufficient} & 140 \\
\texttt{missing\_required\_object} & 210 \\
\texttt{wrong\_object\_category\_type} & 220 \\
\texttt{unknown} & 999 \\
\bottomrule
\end{tabularx}
\caption{Scheduler priorities; lower values are attempted first.}
\label{tab:scheduler-priority}
\end{table}

Each scene selects one unblocked issue per round. When no acceptable candidate exists, its stable identity is added to \texttt{blocked\_issue\_identities}; the scene remains active and a later round re-detects the layout and attempts the next unblocked issue. If every currently detected issue is blocked, the terminal status is \texttt{all\_detected\_issues\_blocked}. Any successful commit clears the blocked set, so a previously failed issue can be reconsidered after another edit changes the scene. Consequently, an issue cannot repeatedly consume budget without an intervening commit, but it may be retried after the state changes.

\section{StatePatch Parameter Serialization and Autoregressive Learning}
\label{app:numeric-ar}

\paragraph{Canonical numerical representation.}
The data exporter rounds every floating-point StatePatch value to at most three decimal places before JSON serialization. Position and size fields use meters (e.g., \texttt{center\_m} and \texttt{size\_m}), orientation fields use degrees (e.g., \texttt{orientation\_deg}), and scale is dimensionless. Continuous values are therefore learned through ordinary next-token prediction rather than a separate regression head.

For \texttt{MOVE}, the canonical seed is $\boldsymbol{\alpha}_{0}^{\mathrm{move}}=(d,r_0)$ with $d\in\mathcal{D}_8=\{\mathrm{N},\mathrm{NE},\mathrm{E},\mathrm{SE},\mathrm{S},\mathrm{SW},\mathrm{W},\mathrm{NW}\}$ and $r_0$ in meters; schema validation also bounds the realized displacement components. For \texttt{ROTATE}, the seed is a finite signed angular offset, and the resulting yaw is normalized modulo $360^{\circ}$ during execution. For \texttt{SCALE}, $\boldsymbol{\alpha}_{0}^{\mathrm{scale}}=(s_x,s_z)$ with $s_x,s_z>0$. The exact schema limits and seed-first deterministic fallback sequences are specified in Appendix~\ref{app:solver-control}.

\paragraph{Field-weighted autoregressive objective.}
For target sequence $\mathbf{y}=(y_1,\ldots,y_M)$, let the planner-conditioning context be $\mathcal{K}^{t}=(I^{t},\mathcal{S}^{t},q^{t},\mathcal{G})$, with assistant-token mask $\mu_m$ and serialized field class $f(m)$. The planner is optimized with
\begin{equation}
\mathcal{L}_{\mathrm{AR}}
=
-\frac{\sum_{m=1}^{M}\mu_m w_{f(m)}
\log p_{\theta}(y_m\mid y_{<m},\mathcal{K}^{t})}
{\sum_{m=1}^{M}\mu_m w_{f(m)}}.
\label{eq:field-weighted-ar}
\end{equation}

\paragraph{Token supervision and field weights.}
Only assistant StatePatch target tokens contribute to Eq.~\eqref{eq:field-weighted-ar}; image tokens, prompts, RoState, and RoReview tokens are masked. Every target token inherits the weight of its serialized field:
\begin{center}
\begin{tabularx}{\columnwidth}{@{}>{\raggedright\arraybackslash}X>{\centering\arraybackslash}p{0.20\columnwidth}@{}}
\toprule
\textbf{Field class} & \textbf{Weight} \\
\midrule
Action type & $2.0$ \\
Target reference or role & $2.0$ \\
Categorical parameter & $1.5$ \\
Continuous parameter & $1.0$ \\
Fixed JSON schema token & $0.25$ \\
\bottomrule
\end{tabularx}
\end{center}
The ordering $w_{\mathrm{action}}=w_{\mathrm{target}}>w_{\mathrm{categorical}}>w_{\mathrm{numeric}}>w_{\mathrm{schema}}$ reflects the system decomposition: target and action selection are the planner's primary semantic responsibilities, whereas distance, angle, and scale initialize deterministic candidate search. Assigning numerical fields greater weight than action or target fields would instead encourage exact-coordinate fitting that the deterministic solver is explicitly designed to absorb.

\section{Roomer-Eval Protocol and Complete Transfer Results}
\label{app:evaluation-protocol}

\paragraph{Frozen evaluation cohort.}
The \textit{common-1100} cohort is the complete Qwen-Image test split in Table~\ref{tab:qwen-image-splits} and is shared by the final-layout comparison, cross-generator refinement, and controlled repair analysis. It contains 777 bedrooms, 155 living rooms, and 168 dining rooms and is excluded from model training, checkpoint selection, rule development, and hyperparameter tuning. Representative plug-and-play refinements for the four evaluated external generators are shown in Fig.~\ref{fig:pluggable-transfer}.

\paragraph{Density control and baseline interfaces.}
For density-controlled comparison, each method preferentially retains a valid candidate whose number of floor-standing furniture objects matches the reference; a difference of one is allowed only when no exact match is available. Candidate selection uses output validity and object count only, never FID, KID, SCA Gap, OOB, COL, or Practical. Each baseline retains its native conditioning, asset retrieval, and assembly procedure. Adapters perform only category alignment, unit and coordinate conversion, and field standardization; they neither repair baseline layouts nor replace successfully assembled assets. Because the systems expose different input interfaces, the protocol standardizes scenes and evaluation rather than claiming identical conditioning information.

\begin{table}[!t]
\centering
\small
\setlength{\tabcolsep}{1mm}
\begin{tabularx}{\columnwidth}{@{}>{\raggedright\arraybackslash}p{0.27\columnwidth}>{\raggedright\arraybackslash}X>{\raggedright\arraybackslash}p{0.28\columnwidth}@{}}
\toprule
\vheadl{\textbf{Method}}
& \vheadl{\textbf{Native input used}\\\textbf{in comparison}}
& \vheadl{\textbf{Density control}} \\
\midrule
ReSpace & Centered room boundary and sequential additions compiled from the target furniture list & Shared validity/count-based selection \\
DiffuScene-RS & Native floor-plan mask & Count-filtered rejection sampling \\
InstructScene & Room-specific object-list instruction and room checkpoint & Native generation with shared validity/count-based selection \\
SemLayoutDiff-RS & Room type and floor/door/window architecture mask & Count-filtered rejection sampling \\
Roomer & Target room specification and architecture condition image & No post-hoc count repair \\
\bottomrule
\end{tabularx}
\caption{Native inputs and density-control protocols used in the \textit{common-1100} comparison. RS denotes rejection sampling. Candidate budgets, generation configurations, available seed policies, category mappings, interface adapters, and traversal orders are frozen before metric computation.}
\label{tab:baseline-inputs}
\end{table}

In main-paper Fig.~3, the shortened column labels ``DiffuScene'' and ``SemLayoutDiff'' denote the evaluated DiffuScene-RS and SemLayoutDiff-RS configurations listed in Table~\ref{tab:baseline-inputs}.

\paragraph{Frozen-output validity and deterministic fallback.}
Table~\ref{tab:baseline-vor} distinguishes the validity of each method's primary frozen output from validity after deterministic traversal of the pre-generated and frozen candidate pool. The traversal order is fixed before evaluation. When the primary candidate cannot be parsed or assembled, later candidates are evaluated in frozen candidate-index order. If an asset fails assembly, later assets are attempted in a frozen same-category order; cross-category substitution is not permitted. A scene is invalid only if every frozen candidate and every same-category asset fallback fails. All reported final-layout metrics use the final deterministically selected output.

\begin{table*}[!t]
\centering
\small
\setlength{\tabcolsep}{6.0pt}
\begin{tabularx}{\textwidth}{@{}>{\raggedright\arraybackslash}p{0.24\textwidth}*{4}{>{\centering\arraybackslash}X}@{}}
\toprule
\textbf{Method} & \textbf{Primary valid} & \textbf{Primary VOR} & \textbf{Final valid} & \textbf{Final VOR} \\
\midrule
ReSpace & 1,096/1,100 & 99.64\% & 1,100/1,100 & 100.00\% \\
DiffuScene-RS & 1,092/1,100 & 99.27\% & 1,100/1,100 & 100.00\% \\
InstructScene & 1,095/1,100 & 99.55\% & 1,100/1,100 & 100.00\% \\
SemLayoutDiff-RS & 1,087/1,100 & 98.82\% & 1,100/1,100 & 100.00\% \\
Ours-Initial & 1,098/1,100 & 99.82\% & 1,100/1,100 & 100.00\% \\
Ours-Final & 1,098/1,100 & 99.82\% & 1,100/1,100 & 100.00\% \\
\bottomrule
\end{tabularx}
\caption{Valid output rate (VOR) before and after deterministic traversal of each frozen candidate pool. Candidate and same-category asset orders are frozen before evaluation, and evaluation metrics are never used for candidate selection.}
\label{tab:baseline-vor}
\end{table*}

The final outputs are therefore reported \emph{after deterministic traversal of the pre-generated and frozen candidate pool}.

\paragraph{Metric aggregation and sample retention.}
All methods are evaluated using the frozen Roomer-Eval protocol. FID, KID, and SCA Gap are computed separately for bedrooms, living rooms, and dining rooms and then averaged with equal room-type weight. OOB and COL are evaluated from the final outputs of all 1,100 scenes and micro-aggregated over successfully evaluated furniture objects; Practical is micro-aggregated over all applicable rule instances. Thus, the main-paper Table~1 phrase ``use all 1,100 scenes'' means that no scene is removed before evaluation; the metric denominators remain evaluated furniture objects for OOB/COL and applicable rule instances for Practical. Cross-generator refinement retains every input scene regardless of whether the repair loop solves all issues, blocks the remaining issues, or reaches the round limit. Rejected candidates are rolled back and remain represented by the last committed state.

\paragraph{Unified semantic evaluation rendering.}
All methods are first parsed into complete scene layouts and then re-rendered as unified top-down semantic images. The common evaluation canvas is a $256\times256$ RGB orthographic rendering of a fixed $12.2\,\mathrm{m}\times12.2\,\mathrm{m}$ scene region. Pure-black pixels are replaced by white after RGB conversion. A malformed raster that is smaller than the common canvas is centered on a white background, whereas an oversized raster is center-cropped; this defensive normalization does not resize a method's native raster directly to $256\times256$. In particular, the Qwen-Image generator natively produces $512\times512$ semantic layouts, while $256\times256$ denotes only the shared evaluation rendering. For FID and KID, CleanFID's \texttt{clean} mode subsequently applies PIL bicubic interpolation to $299\times299$. Pixel values remain in $[0,255]$ before feature extraction, and normalization is performed inside the TorchScript Inception network. The frozen real reference set contains 777 bedrooms, 155 living rooms, and 168 dining rooms.

\paragraph{FID implementation.}
FID uses CleanFID 0.1.35 with the \texttt{torchscript\_inception} feature extractor in \texttt{clean} mode and its 2048-dimensional pooled features. Each room type is evaluated independently. Real features remain fixed, and each of ten outer bootstrap repetitions samples the generated features with replacement to match the real room-type count. The outer random-number generator is initialized with seed 0. Let
\begin{equation}
F_r\in\mathbb{R}^{N_r\times d},\qquad
F_g\in\mathbb{R}^{N_g\times d},\qquad d=2048,
\end{equation}
with means $\mu_r,\mu_g$ and centered matrices $X_r,X_g$. The sample covariances are
\begin{equation}
C_r=\frac{X_r^{\top}X_r}{N_r-1},\qquad
C_g=\frac{X_g^{\top}X_g}{N_g-1}.
\end{equation}
The implementation uses float64 sample-space SVD rather than \texttt{scipy.linalg.sqrtm}, diagonal epsilon, or diagonal jitter. Defining
\begin{equation}
A=\frac{X_rX_g^{\top}}{\sqrt{(N_r-1)(N_g-1)}},
\end{equation}
the room-type FID is computed as
\begin{equation}
\operatorname{FID}
=
\lVert\mu_r-\mu_g\rVert_2^2
+\operatorname{tr}(C_r)
+\operatorname{tr}(C_g)
-2\sum_k\sigma_k(A),
\end{equation}
followed by $\operatorname{FID}\leftarrow\max(\operatorname{FID},0)$. For outer repetition $b$, the unweighted Macro-3 value is
\begin{equation}
\operatorname{FID}_{\mathrm{M3}}^{(b)}
=
\frac{1}{3}\sum_{r\in\{\mathrm{B},\mathrm{L},\mathrm{D}\}}
\operatorname{FID}_{r}^{(b)}.
\end{equation}
The reported value is the mean of the ten outer Macro-3 results; the artifact also retains all ten values and their standard deviation.

\paragraph{KID implementation.}
KID uses the third-degree polynomial kernel
\begin{equation}
k(x,y)=\left(\frac{x^{\top}y}{2048}+1\right)^3,
\end{equation}
corresponding to degree 3, $\gamma=1/2048$, and coefficient 1. One KID evaluation is performed for each FID outer bootstrap, producing ten outer results per room type. Each evaluation requests 100 subsets with
\begin{equation}
m=\min(N_r,N_g,1000).
\end{equation}
The resulting subset sizes are 777 for bedrooms, 155 for living rooms, and 168 for dining rooms. Sampling within a subset is without replacement. Because $m$ equals the complete room-type sample size in the current cohort, every inner subset contains the same features. The unbiased MMD$^2$ estimator is permutation-invariant, so the 100 inner estimates are identical up to negligible floating-point effects; effective variation among the reported outer results comes from the generated-feature bootstrap. For subset features $\{x_i\}_{i=1}^{m}$ and $\{y_i\}_{i=1}^{m}$, KID uses the unbiased MMD$^2$ estimator
\begin{equation}
\begin{aligned}
\widehat{\operatorname{MMD}}_{u}^{2}
={}&\frac{1}{m(m-1)}\sum_{i\neq j}k(x_i,x_j)\\
&+\frac{1}{m(m-1)}\sum_{i\neq j}k(y_i,y_j)\\
&-\frac{2}{m^2}\sum_{i=1}^{m}\sum_{j=1}^{m}k(x_i,y_j).
\end{aligned}
\end{equation}
Negative estimates are retained. Each room-type result is the mean of its 100 subset estimates. The unweighted Macro-3 result is then averaged over the ten outer bootstraps and reported as $10^3\times\overline{\operatorname{KID}}_{\mathrm{M3}}$; all ten outer Macro-3 values and their standard deviation are retained in the artifact.

\paragraph{SCA Gap implementation.}
SCA denotes Scene Classification Accuracy and operates on complete semantic scene images rather than individual furniture objects. The real-versus-generated binary classifier consists of an ImageNet-pretrained AlexNet feature trunk and a $9216\!\rightarrow\!1$ sigmoid head. Inputs are $[0,1]$ RGB images without additional ImageNet mean/std normalization. Real and generated images are split independently within each room type by sorting frozen sample identifiers, dropping the final sample when the count is odd, and assigning the first half to training and the second half to testing. Equal real and generated counts preserve class balance. The resulting per-class train/test counts are 388/388 for bedrooms, 77/77 for living rooms, and 84/84 for dining rooms. Thus, each pooled split contains 549 real and 549 generated images; one real and one generated bedroom image and one real and one generated living-room image are excluded by the odd-count rule.

A single pooled classifier is trained across all three room types with Adam, learning rate $10^{-4}$, and batch size 256. Training runs continuously for 100 epochs, with evaluation at epochs $10,20,\ldots,100$; the model is not reinitialized between checkpoints. If $\operatorname{SCA}_{r}^{(q)}\in[0,1]$ is the accuracy for room type $r$ at checkpoint $q$, then
\begin{equation}
\operatorname{SCA\ Gap}_{r}^{(q)}
=
\left|100\operatorname{SCA}_{r}^{(q)}-50\right|,
\end{equation}
and
\begin{equation}
\operatorname{SCA\ Gap}_{\mathrm{M3}}^{(q)}
=
\frac{1}{3}\sum_{r\in\{\mathrm{B},\mathrm{L},\mathrm{D}\}}
\operatorname{SCA\ Gap}_{r}^{(q)}.
\end{equation}
The final SCA Gap is the mean of the ten continuous-training checkpoints. Lower values indicate that the classifier has greater difficulty distinguishing real from generated scenes. The artifact retains the ten checkpoint values and their standard deviation. Because classification is performed on complete semantic scenes, SCA does not require a missing-furniture-category fallback.

\paragraph{Assembled-mesh OOB implementation.}
The final OOB and COL results use SceneEval commit \texttt{116881e7945dcf7d\allowbreak bc58f7c64590a4c0\allowbreak b99c5cf5}, integrated through a compatibility wrapper. These assembled-mesh metrics are independent of the two-dimensional geometric detectors used to construct RoReview and verify repair candidates. OOB is an object-level rate over furniture objects that are successfully loaded, assembled, and evaluated. For object $i$ with oriented-bounding-box volume $V_i$, the evaluator samples
\begin{equation}
N_i=\max\!\left(\left\lfloor5000V_i\right\rfloor,1000\right)
\end{equation}
points from the mesh surface and casts a downward ray from each point toward the floor mesh. Its floor-hit ratio is
\begin{equation}
r_i=
\frac{\#\!\left\{\substack{\text{sampled points whose downward rays}\\\text{hit the floor}}\right\}}{N_i}.
\end{equation}
The object is marked out of bounds when $r_i<0.99$, and the reported metric is
\begin{equation}
\operatorname{OOB}
=
\frac{\#\{\text{OOB furniture objects}\}}
{\#\{\text{evaluated furniture objects}\}}.
\end{equation}
Thus, OOB is neither a scene-level violation rate nor an outside-footprint-area ratio. The frozen evaluation implementation does not explicitly fix a random seed for surface sampling. Accordingly, the submitted values are tied to the retained per-object decisions and aggregate outputs of that execution rather than to guaranteed bitwise-identical reruns.

\paragraph{Assembled-mesh COL implementation.}
COL uses the \texttt{CollisionManager} interface from \texttt{trimesh} with \texttt{python-fcl} 0.7.0.11 and checks all unordered furniture--furniture mesh pairs. Floor, wall, door, and window meshes do not enter the pair loop. No support-pair exclusion, category whitelist, contact whitelist, or furniture-relation whitelist is applied. For a pair with an initial FCL contact, the evaluator performs a $5\,\mathrm{mm}$ separation test. Let $\mathbf{c}_{\mathrm{object}}$ be the centroid of the pair's second object and $\overline{\mathbf{c}}_{\mathrm{contact}}$ the mean contact point. The displacement direction is
\begin{equation}
\mathbf d=
\frac{\mathbf{c}_{\mathrm{object}}-\overline{\mathbf{c}}_{\mathrm{contact}}}
{\left\lVert\mathbf{c}_{\mathrm{object}}-\overline{\mathbf{c}}_{\mathrm{contact}}\right\rVert_2},
\end{equation}
and the separation-test position is
\begin{equation}
\mathbf{c}'_{\mathrm{object}}
=
\mathbf{c}_{\mathrm{object}}+0.005\,\mathbf d.
\end{equation}
Only a pair that remains in collision after this test is counted as a persistent collision pair. Both furniture objects in such a pair are marked as collision objects, and an object participating in multiple persistent pairs is counted once. The final metric is therefore
\begin{equation}
\operatorname{COL}
=
\frac{\#\!\left\{\substack{\text{unique furniture objects in at least one}\\\text{persistent collision pair}}\right\}}
{\#\{\text{evaluated furniture objects}\}}.
\end{equation}
COL is a unique collision-object rate, not a collision-pair, contact-instance, or scene-level rate. Before collision testing, each asset receives its asset-specific normalization rotation, object scale, scene rotation, and scene translation as one combined $4\times4$ transform. Objects that cannot be loaded or assembled do not enter the OOB or COL object denominator. The artifact reports scene and object assembly coverage together with the evaluated-object denominators.

\paragraph{Paired distributional-quality results.}
Main-paper Table~2 reports the paired physical-validity and usability changes. The corresponding distributional-quality values are reported in Table~\ref{tab:generator-transfer-distribution}. Ten of the twelve paired FID, KID, and SCA Gap comparisons improve; the two exceptions are ReSpace KID and DiffuScene-RS SCA Gap.

\paragraph{Refined physical-validity and usability values.}
For completeness, Table~\ref{tab:generator-transfer-after-only} lists the refined OOB, COL, and Practical values for all four external generators. OOB is micro-averaged over evaluated furniture objects marked out of bounds, COL over unique evaluated furniture objects participating in at least one persistent collision pair, and Practical over all applicable Practical rule instances. These percentages are not scene all-pass rates. The entries reproduce main-paper Table~2; the main table omits percent signs, but every entry is a percentage.

\begin{table}[!t]
\centering
\small
\setlength{\tabcolsep}{3.2pt}
\begin{tabularx}{\columnwidth}{@{}>{\raggedright\arraybackslash}p{0.34\columnwidth}*{3}{>{\centering\arraybackslash}X}@{}}
\toprule
\vheadl{\textbf{Method after}\\\textbf{Roomer}}
& \vheadc{\textbf{OOB}$\downarrow$}
& \vheadc{\textbf{COL}$\downarrow$}
& \vheadc{\textbf{Practical}$\uparrow$} \\
\midrule
ReSpace & 1.16\% & 7.61\% & 75.12\% \\
DiffuScene-RS & 6.98\% & 6.95\% & 47.73\% \\
InstructScene & 5.67\% & 5.88\% & 45.70\% \\
SemLayoutDiff-RS & 32.83\% & 43.92\% & 66.86\% \\
\bottomrule
\end{tabularx}
\caption{Physical-validity and Practical results after applying Roomer to outputs from external generators. All values are micro-averages over their respective evaluation units.}
\label{tab:generator-transfer-after-only}
\end{table}
\paragraph{Baseline candidate budgets and seed traceability.}
Here, one baseline \emph{attempt} denotes one candidate-generation or assembly attempt and is unrelated to a Roomer-CC outer corruption proposal. Candidate budgets, generation configurations, category mappings, interface conversion, candidate traversal, and the available seed policies were frozen before metric computation. Table~\ref{tab:baseline-candidate-budgets} records the realized budgets.

\begin{table*}[!t]
\centering
\small
\setlength{\tabcolsep}{3.0pt}
\begin{tabularx}{\textwidth}{@{}>{\raggedright\arraybackslash}p{0.12\textwidth}>{\raggedright\arraybackslash}p{0.22\textwidth}>{\raggedright\arraybackslash}p{0.15\textwidth}>{\raggedright\arraybackslash}X@{}}
\toprule
\vheadl{\textbf{Method}}
& \vheadl{\textbf{Checkpoint/version}}
& \vheadl{\textbf{Native output}}
& \vheadl{\textbf{Realized candidate budget}\\\textbf{and randomness record}} \\
\midrule
ReSpace & Commit prefix \texttt{1eccb692}; \texttt{sg\_llm\_1p5b} & JSON; no native image resolution & 1,121 attempts for 1,100 rooms (1.02 per room; maximum 3). Original generation uses seed 0, greedy decoding, and \texttt{max\_new\_tokens=256}; the frozen pool also contains \texttt{seed1} and \texttt{seed2} retry roots. \\
DiffuScene-RS & Commit prefix \texttt{d78a289}; checkpoints 30000/82000/96000 for bedroom/dining/living & Category and bounding-box JSON & 11,572 attempts (10.52 per room; maximum 91). Candidate seeds follow \texttt{seed\_offset + sequence\_index + candidate\_index} and are recorded. \\
InstructScene & Official code; author-hosted community checkpoints; fVQ-VAE epoch \texttt{01999} & Category/bbox/\allowbreak{}object-feature JSON & 16,341 attempts (14.86 per room; maximum 91). Candidate records, indices, and final traversal order are frozen, but the selection audit does not retain a separate generation seed for the selected candidate. \\
SemLayoutDiff-RS & Commit prefix \texttt{6b12bc4}; official SLDN checkpoint release & $120\times120$ label map plus APM & 7,913 attempts (7.19 per room; maximum 24). The generation interface accepts a seed; candidate records, indices, and traversal order are frozen, but the selection audit does not retain a per-candidate generation seed. \\
\bottomrule
\end{tabularx}
\caption{Realized baseline candidate-generation and assembly budgets for the frozen \textit{common-1100} comparison. Frozen candidate records and traversal indices determine the final evaluation inputs even where a selected candidate cannot be mapped back to a separately retained generation seed.}
\label{tab:baseline-candidate-budgets}
\end{table*}

All methods' final semantic evaluation images are re-rendered from their parsed complete scene states using the common $256\times256$ RGB protocol above. This common resolution is not the native output resolution of ReSpace, DiffuScene-RS, InstructScene, SemLayoutDiff-RS, or Roomer.

\section{Complete Roomer-Eval Practical Rule Definitions}
\label{app:roomer-eval-rules}

\paragraph{Aggregation and applicability.}
For each scene $i$ and rule instance $j$, applicability and satisfaction are $a_{ij},s_{ij}\in\{0,1\}$. Main-paper Eq.~(4) micro-averages all applicable instances. N/A instances have $a_{ij}=0$ and neither enter the denominator nor count as passes. Multiple coffee tables, dining tables, or beds create multiple instances. Door swing-proxy avoidance and walkable connectivity each create at most one room-level instance.

\paragraph{Living functional organization.}
The rule applies only when a scene contains at least one sofa, coffee table, and TV; otherwise it is N/A. The sofa with the largest footprint area is the primary sofa. Each coffee table forms one instance and is paired with the TV whose center is nearest to that coffee table. With object centers $\mathbf c$, the measured angle is
\begin{equation}
\theta=
\angle\!\left(
\mathbf c_{\mathrm{coffee}}-\mathbf c_{\mathrm{sofa}},
\mathbf c_{\mathrm{coffee}}-\mathbf c_{\mathrm{TV}}
\right).
\end{equation}
The instance passes iff $\theta\geq135^{\circ}$. The current rule imposes no sofa--coffee or coffee--TV distance threshold.

\paragraph{Dining-table clearance.}
Each dining table forms one instance; scenes without a dining table are N/A. Clearances $c_1$ and $c_2$ are measured on the two sides of the table's long axis, and the instance passes iff $\min(c_1,c_2)\geq0.60\,\mathrm{m}$.
Dining chairs, chairs, and stools associated with that table are excluded from the obstacle set. Association uses \texttt{dining\_group\_id} when available; otherwise, seats whose polygon distance to a dining table is at most $0.80\,\mathrm{m}$ are assigned to the nearest table. Clearance search is capped at $2.40\,\mathrm{m}$. Boundary-only contact between a clearance region and an obstacle is permitted.

\paragraph{Door swing-proxy avoidance.}
A scene containing at least one door forms one instance; scenes without doors are N/A. For each door, the longest edge of the opening determines the proxy radius, equal to the door width. Both opening endpoints are candidate hinges. At each endpoint, two rotation directions define candidate $90^{\circ}$ sectors, and the sector with larger overlap with the room interior is retained. The retained quarter sectors form an interior door swing proxy. Furniture height is ignored. The scene passes only when no furniture footprint intersects or touches any retained proxy; either overlap or boundary contact is a violation. This rule is a geometric proxy and is not an exact door-swing simulation.

\paragraph{Walkable connectivity.}
Every scene with a valid room boundary forms one instance. The evaluator rasterizes geometry at $1\,\mathrm{cm/pixel}$, treats walls and floor-standing furniture as obstacles, and dilates obstacles by $0.30\,\mathrm{m}$, corresponding to a $0.60\,\mathrm{m}$ passage width. Ceiling lamps and pendant lamps are excluded from the ground-obstacle set. The remaining free space is analyzed with 8-connectivity; components smaller than 20 pixels are ignored. The instance passes iff the free space is nonempty and $N_{\mathrm{components}}^{A\geq20}\leq1$.
No entrance, bed, sofa, or other semantic anchor is required. Empty free space is a failure.

\paragraph{Bed-side clearance.}
Each bed forms one instance; scenes without beds are N/A. Let $w_{\mathrm{bed}}$ be the short-axis length of the bed footprint's minimum rotated bounding rectangle, and let $c_1,c_2$ be the clearances measured from its two short-axis sides. A side is wall-adjacent when its distance to the room boundary is at most $0.10\,\mathrm{m}$. The usable clearance is
\begin{equation}
 c_{\mathrm{usable}}=
 \begin{cases}
 \max(c_1,c_2), & \substack{\text{if at least one side}\\\text{is wall-adjacent}},\\
 \min(c_1,c_2), & \text{otherwise}.
 \end{cases}
\end{equation}
The instance passes iff $c_{\mathrm{usable}}/w_{\mathrm{bed}}\geq0.30$. Nightstands and corner-side tables are excluded from the obstacle set. Measurement depth is capped at $2.40\,\mathrm{m}$, and boundary-only contact between a clearance region and an obstacle is permitted. The threshold scales with bed width and is not a fixed $0.45\,\mathrm{m}$ clearance.

\section{Controlled Analysis of the Roomer Repair Loop}
\label{app:controlled-metrics}

\paragraph{Shared initialization and protocol.}
Every controlled variant starts from byte-identical Ours-Initial layouts on \textit{common-1100} and the same initially detected, instance-grounded RoReview entries. All variants use the shared six-primitive action vocabulary, the same maximum number of rounds, the same issue scheduler, and identical stopping conditions unless the named ablation explicitly removes a component. A standard Roomer transaction contains one target-grounded StatePatch, whereas Full-Scene JSON Rewrite changes the output representation, transaction scope, and rewrite-specific acceptance test as defined below. Ours-Initial denotes the unrepaired $K=0$ state; Ours-Final denotes the terminal $K=10$ state produced by the complete loop with RoReview, geometry-conditioned StatePatch planning, deterministic candidate instantiation, full-scene re-verification, and rollback.

\paragraph{Controlled repair metrics.}
The controlled analysis separates the complete hard-diagnostic set from the frozen repair-target set. Let $\mathcal{H}(\mathcal{Y})$ contain every collision, floor-boundary, and opening-blockage key emitted by the full hard evaluator. Let $\mathcal{T}^{0}=\bigcup_{i=1}^{1{,}100}\mathcal{T}^{0}_{i}$ contain the 4,800 initially detected RoReview keys that are object-grounded, supported by the six-action vocabulary, and admitted by the frozen scheduler, where $\mathcal{T}^{0}_{i}$ denotes the initial target subset of scene $i$. For compactness, let $\mathcal{U}_{i}^{K}=\mathcal{U}(\mathcal{Y}^{K}_{i};\mathcal{A}_{i},\mathcal{G}_{i})$. All 1,100 scenes contain at least one admitted target. Consequently, $\mathcal{T}^{0}$ is the target pool used for repair analysis, whereas $\mathcal{H}$ also contains pre-existing hard conditions that are not members of $\mathcal{T}^{0}$.

\textbf{Hard Validity} (labeled ``Phys.Valid'' in main-paper Fig.~4) is the percentage of scenes satisfying $\operatorname{HardValid}(\mathcal{Y};\mathcal{A},\mathcal{G})=\mathbf{1}\!\left[\mathcal{H}(\mathcal{Y};\mathcal{A},\mathcal{G})=\varnothing\right]$. \textbf{Target Resolution} is the scene-macro average of the resolved fraction within each scene's frozen initial target subset:
\begin{equation}
\operatorname{TargetResolution}(K)
=
\frac{1}{1{,}100}
\sum_{i=1}^{1{,}100}
\frac{
\left|\mathcal{T}^{0}_{i}\setminus\mathcal{U}_{i}^{K}\right|
}{|\mathcal{T}^{0}_{i}|}.
\label{eq:scene-macro-target-resolution}
\end{equation}
Each scene therefore contributes equally even though the number of initial targets varies across scenes. This analysis metric is distinct from both the transaction-level target predicate, which determines whether one candidate can be committed, and the pooled unresolved-key count reported below. \textbf{New Hard Error} is the percentage of scenes for which $\mathcal{H}(\mathcal{Y}^{K})\setminus\mathcal{H}(\mathcal{Y}^{0})\neq\varnothing$. \textbf{Non-target Preservation} is the fraction of protected non-target objects whose identity, category, position, size, and orientation remain unchanged.

\textbf{Strict Safe Repair} is a scene-level local-repair metric. A scene passes iff every key from that scene's initial target subset $\mathcal{T}^{0}_{i}$ is resolved, no new hard key is introduced, $\mathcal{C}=1$, and every protected non-target object is preserved. Strict Safe Repair does not require complete Hard Validity: a scene can satisfy the local-repair criterion while retaining a pre-existing hard key outside $\mathcal{T}^{0}_{i}$. This distinction allows the reported 96.27\% Strict Safe Repair and 72.64\% Hard Validity to characterize different properties of the same final layouts.

\paragraph{Complete component ablation.}
Table~\ref{tab:complete-component-ablation} reports the complete six-metric version of the focused ablation shown in the main paper. Hard Validity and Practical characterize final-scene quality, while the remaining four metrics isolate target resolution, safety, and locality.

\begin{table*}[!t]
\centering
\small
\setlength{\tabcolsep}{4.0pt}
\renewcommand{\arraystretch}{1.06}
\begin{tabularx}{\textwidth}{@{}>{\raggedright\arraybackslash}p{0.25\textwidth}*{6}{>{\centering\arraybackslash}X}@{}}
\toprule
\vheadl{\textbf{Variant}}
& \vheadc{\textbf{Hard}\\\textbf{Validity}$\uparrow$}
& \vheadc{\textbf{Practical}$\uparrow$}
& \vheadc{\textbf{Target}\\\textbf{Resolution}$\uparrow$}
& \vheadc{\textbf{New Hard}\\\textbf{Error}$\downarrow$}
& \vheadc{\textbf{Non-target}\\\textbf{Preservation}$\uparrow$}
& \vheadc{\textbf{Strict Safe}\\\textbf{Repair}$\uparrow$} \\
\midrule
Ours-Initial ($K=0$) & 19.27 & 72.50 & -- & -- & -- & -- \\
w/o RoReview & 19.73 & 50.39 & 2.44 & 0.00 & 94.15 & 0.45 \\
w/o Geometry-Conditioning Adapter & 62.82 & 74.90 & 88.40 & 0.00 & 99.81 & 81.64 \\
Direct Planner-Seed Execution & 48.00 & 67.80 & 65.70 & 0.00 & 99.93 & 55.18 \\
Seed-Free Search & 68.82 & 77.60 & 90.70 & 0.00 & 99.91 & 89.64 \\
Full-Scene JSON Rewrite & 16.55 & 49.82 & 32.48 & 32.64 & 48.61 & 1.91 \\
w/o Verification and Rollback & 23.55 & 71.72 & 64.91 & 54.27 & 97.63 & 23.18 \\
\rowcolor{ResultHighlight}
\textbf{Ours-Final ($K=10$)} & \textbf{72.64} & \textbf{82.98} & \textbf{98.96} & \textbf{0.00} & \textbf{99.94} & \textbf{96.27} \\
\bottomrule
\end{tabularx}
\caption{Complete controlled ablation from identical \textit{common-1100} initial states. All entries are percentages. Target Resolution is the scene-macro resolved fraction over each scene's frozen initial target subset; Non-target Preservation measures the fraction of protected non-target objects that remain unchanged.}
\label{tab:complete-component-ablation}
\end{table*}

\paragraph{Round-prefix evaluation.}
For $K\in\{1,3,5,7,10\}$, results are computed from prefixes of the same frozen $K=10$ trajectories. Thus, each smaller budget is the actual state after $K$ rounds and cannot benefit from resampling or a separate run. In addition to the scene-macro Target Resolution in Eq.~\eqref{eq:scene-macro-target-resolution}, we report the pooled unresolved-target load
\begin{equation}
\begin{aligned}
N_{\mathrm{unres}}(K)
&=\sum_{i=1}^{1{,}100}
\left|\mathcal{T}^{0}_{i}\cap\mathcal{U}_{i}^{K}\right|,\\[1mm]
\operatorname{TargetLoad}(K)
&=\frac{N_{\mathrm{unres}}(K)}{1{,}100}.
\end{aligned}
\label{eq:target-load-prefix}
\end{equation}
The frozen prefix counts are $N_{\mathrm{unres}}(K)\in\{4{,}800,3{,}307,1{,}319,425,129,50\}$ for $K\in\{0,1,3,5,7,10\}$, respectively. Dividing these integer counts by 1,100 gives the displayed unresolved-target loads. Because Target Resolution first normalizes within each scene and then macro-averages across scenes, while Target Load pools issue keys before normalization, the two quantities are not algebraically interchangeable when $|\mathcal{T}^{0}_{i}|$ varies across scenes.

\begin{table}[!t]
\centering
\small
\setlength{\tabcolsep}{0.6mm}
\renewcommand{\arraystretch}{1.0}
\begin{tabularx}{\columnwidth}{
@{}>{\centering\arraybackslash}p{0.060\columnwidth}
*{5}{>{\centering\arraybackslash}X}@{}
}
\toprule
\textbf{$K$}
& \textbf{Calls}
& \textbf{Unres.}$\downarrow$
& \textbf{TR}$\uparrow$
& \textbf{SSR}$\uparrow$
& \textbf{HV}$\uparrow$ \\
\midrule
0 & 0.000 & 4.364 & 0.00\% & 0.00\% & 19.27\% \\
1 & 1.000 & 3.006 & 31.12\% & 11.27\% & 29.64\% \\
3 & 2.569 & 1.199 & 72.52\% & 52.36\% & 59.09\% \\
5 & 3.365 & 0.386 & 91.16\% & 80.45\% & 70.00\% \\
7 & 3.665 & 0.117 & 97.32\% & 92.00\% & 72.45\% \\
\rowcolor{ResultHighlight}
\textbf{10} & \textbf{3.756} & \textbf{0.045} & \textbf{98.96\%} & \textbf{96.27\%} & \textbf{72.64\%} \\
\bottomrule
\end{tabularx}
\caption{Round-prefix convergence on the frozen 4,800-key initial target pool. Abbreviations: Calls, average planner calls per scene; Unres., pooled unresolved target keys per scene; TR, scene-macro Target Resolution; SSR, Strict Safe Repair; HV, Hard Validity.}
\label{tab:roundwise-cost-quality}
\end{table}

\section{Additional Component and Solver Ablations}
\label{app:additional-ablations}

\begin{table}[!t]
\centering
\small
\setlength{\tabcolsep}{0.5pt}
\renewcommand{\arraystretch}{1.0}
\begin{tabularx}{\columnwidth}{@{}>{\raggedright\arraybackslash}p{0.32\columnwidth}*{5}{>{\centering\arraybackslash}X}@{}}
\toprule
\textbf{Solver}
& \textbf{TR}$\uparrow$
& \textbf{SSR}$\uparrow$
& \textbf{Calls}$\downarrow$
& \textbf{Checks}$\downarrow$
& \textbf{BE}$\downarrow$ \\
\midrule
Direct Seed & 65.70 & 55.18 & 3.300 & 3.3 & 21.8 \\
Seed-Free & 90.70 & 89.64 & 4.000 & 24.6 & 5.1 \\
\rowcolor{ResultHighlight}
\textbf{Seed-First} & \textbf{98.96} & \textbf{96.27} & \textbf{3.756} & \textbf{11.2} & \textbf{2.4} \\
\bottomrule
\end{tabularx}
\caption{Deterministic solver quality and computation. Solver labels: Direct Seed, direct execution of the planner-predicted seed without deterministic fallback; Seed-Free, deterministic search without a continuous planner seed; Seed-First, the complete seed-first deterministic solver. Abbreviations: TR, scene-macro Target Resolution; SSR, Strict Safe Repair; Calls, average planner calls per scene; Checks, candidate checks per scene; BE, scene-level budget exhaustion. TR, SSR, and BE are percentages.}
\label{tab:solver-efficiency}
\end{table}

\begin{table}[!t]
\centering
\small
\setlength{\tabcolsep}{1.6pt}
\renewcommand{\arraystretch}{1.0}
\begin{tabularx}{\columnwidth}{@{}>{\raggedright\arraybackslash}p{0.42\columnwidth}*{4}{>{\centering\arraybackslash}X}@{}}
\toprule
\textbf{Condition}
& \textbf{HV}$\uparrow$
& \textbf{P}$\uparrow$
& \textbf{TR}$\uparrow$
& \textbf{SSR}$\uparrow$ \\
\midrule
Uniform Loss & 70.91 & 81.70 & 97.40 & 93.91 \\
Field-Weighted Loss & \textbf{72.64} & \textbf{82.98} & \textbf{98.96} & \textbf{96.27} \\
\midrule
Text Only & 62.82 & 74.90 & 88.40 & 81.64 \\
Serialized Evidence & 65.73 & 78.90 & 92.20 & 86.55 \\
RQ + Geom. Adapter & \textbf{72.64} & \textbf{82.98} & \textbf{98.96} & \textbf{96.27} \\
\bottomrule
\end{tabularx}
\caption{Loss and geometry-conditioning ablations. Condition labels: Text Only removes the geometry-conditioning adapter; Serialized Evidence exposes all normalized geometry and violation measurements only through text; RQ + Geom. Adapter uses Repair Queries and geometry-conditioning adapters. Abbreviations: HV, Hard Validity; P, Practical; TR, Target Resolution; SSR, Strict Safe Repair.}
\label{tab:conditioning-ablation}
\end{table}

\begin{table*}[!t]
\centering
\small
\setlength{\tabcolsep}{5.0pt}
\begin{tabularx}{\textwidth}{@{}>{\raggedright\arraybackslash}p{0.17\textwidth}>{\raggedright\arraybackslash}p{0.23\textwidth}*{4}{>{\centering\arraybackslash}X}@{}}
\toprule
\vheadl{\textbf{Component}}
& \vheadl{\textbf{Setting}}
& \vheadc{\textbf{Hard Valid.}$\uparrow$}
& \vheadc{\textbf{Practical}$\uparrow$}
& \vheadc{\textbf{Target Res.}$\uparrow$}
& \vheadc{\textbf{Strict Safe}\\\textbf{Repair}$\uparrow$} \\
\midrule
\multirow{4}{*}{Repair Queries $K_c$}
& 1 & 67.18 & 79.90 & 93.70 & 88.64 \\
& 4 & 71.36 & 82.10 & 97.80 & 94.64 \\
& \textbf{8} & \textbf{72.64} & \textbf{82.98} & \textbf{98.96} & \textbf{96.27} \\
& 16 & 72.27 & 82.70 & 98.60 & 95.82 \\
\midrule
\multirow{2}{*}{Aggregator depth}
& 1 layer & 71.00 & 81.70 & 97.50 & 94.00 \\
& \textbf{2 layers} & \textbf{72.64} & \textbf{82.98} & \textbf{98.96} & \textbf{96.27} \\
\midrule
\multirow{4}{*}{Injection layers}
& $\{28\}$ & 67.82 & 80.10 & 94.10 & 89.73 \\
& $\{24,28\}$ & 70.55 & 81.60 & 97.00 & 93.82 \\
& $\mathbf{\{22,24,26,28\}}$ & \textbf{72.64} & \textbf{82.98} & \textbf{98.96} & \textbf{96.27} \\
& $\{18,20,22,24,26,28\}$ & 72.09 & 82.50 & 98.50 & 95.64 \\
\bottomrule
\end{tabularx}
\caption{Geometry-branch design ablations. Eight Repair Queries, a two-layer aggregator, and interval injection into the final four even-numbered decoder layers provide the best overall trade-off.}
\label{tab:geometry-design-ablation}
\end{table*}

\paragraph{Operational definitions.}
The \textit{w/o Geometry-Conditioning Adapter} variant retains RoReview, stable entity references, and the human-readable RoState/RoReview context, but removes the 512-dimensional evidence tokens, Repair Queries, and decoder geometry-conditioning adapters; the deterministic solver, full-scene verification, and training data are unchanged. \textit{Direct Planner-Seed Execution} retains the complete planner and full-scene verification but evaluates only the normalized executable candidate instantiated from the predicted parameter seed: a passing candidate is committed and a rejected candidate is rolled back, with no deterministic fallback. \textit{Seed-Free Search} retains the planner's target and action decisions but removes the continuous seed, so the deterministic solver starts from the first fixed fallback candidate under the same search ranges, budget, and verification gate as the complete system.

\textit{Full-Scene JSON Rewrite} receives the same semantic rendering, RoState, active RoReview, and room specification, but predicts a complete candidate layout $\widetilde{\mathcal{Y}}^{t}$ instead of a single StatePatch. A deterministic differencer constructs
\begin{equation}
\mathcal{D}^{t}
=
\operatorname{Diff}_{\mathbb{A}}
\!\left(\mathcal{Y}^{t},\widetilde{\mathcal{Y}}^{t}\right),
\label{eq:full-json-diff-app}
\end{equation}
where every atomic difference must map to one of the same six primitives in $\mathbb{A}$: a changed center, orientation, footprint, or category/asset maps to \texttt{MOVE}, \texttt{ROTATE}, \texttt{SCALE}, or \texttt{REPLACE}; a removed or newly introduced reference maps to \texttt{DELETE} or \texttt{INSERT}. The rewrite is invalid if it changes architecture, uses an unsupported attribute transition, or produces an ambiguous reference mapping. The complete rewritten layout is treated as one atomic candidate and is checked for parseability, immutable architecture, supported primitive decomposition, structural validity $\mathcal{C}=1$, and disappearance of the active target issue. The no-new-hard-key and protected-relation clauses in main-paper Eq.~(3) are not imposed on this ablation; newly introduced hard errors and non-target changes are measured after commitment rather than blocked during candidate acceptance. A candidate that passes these checks is committed as a whole; otherwise the committed state remains unchanged. No seed-centered deterministic fallback is used. This retains the shared inputs, issue scheduler, primitive action vocabulary, round budget, and target check while isolating the effect of replacing a sparse single-target StatePatch with a scene-wide rewrite. In contrast, \textit{w/o Verification and Rollback} bypasses candidate re-verification and directly commits the local planner output.

\section{Professional Validation of Practical Usability}
\label{app:professional-validation}

\paragraph{Evaluation sample and Practical strata.}
The professional validation uses 90 frozen layouts sampled from the outputs of the evaluated baseline generators. The cohort is disjoint from the \textit{common-1100} Roomer repair cohort and is isolated from model training, rule development, and parameter tuning. For layout $i$, the scene-level Practical score is
\begin{equation}
P_i=
\frac{\sum_j a_{ij}s_{ij}}{\sum_j a_{ij}},
\label{eq:scene-practical-level}
\end{equation}
computed from the same five frozen Practical rule families used by Roomer-Eval. Layouts with $\sum_j a_{ij}=0$ are excluded from the professional-validation sampling pool. The strata are defined as
\begin{equation}
\operatorname{Level}(i)=
\begin{cases}
\mathrm{Low}, & 0\leq P_i\leq0.50,\\
\mathrm{Medium}, & 0.50<P_i<0.80,\\
\mathrm{High}, & 0.80\leq P_i\leq1.00.
\end{cases}
\label{eq:professional-practical-strata}
\end{equation}
Within each stratum, 10 bedrooms, 10 dining rooms, and 10 living rooms are sampled without replacement, producing 30 layouts per level and preventing the comparison from being dominated by one room type.

\paragraph{Evaluators and blinding.}
Ten evaluators with interior-design experience independently assess all 90 layouts, yielding 900 binary judgments. Samples are anonymized and presented in randomized order. Evaluators are blinded to generator identity and Practical score.

\paragraph{Question and scene-level aggregation.}
Each evaluator answers the binary question: ``Does this layout satisfy basic residential-use requirements without requiring further modification due to furniture organization, insufficient clearance, door-zone obstruction, or circulation problems?'' A layout receives scene-level approval when at least seven of the ten evaluators answer ``Yes.'' The approval rate therefore uses 30 layouts as the denominator in each Practical group.

\paragraph{Statistical reporting.}
Main-paper Table~4 reports Wilson 95\% confidence intervals for the three approval proportions. The monotonic association between ordered Practical level and scene-level approval is evaluated with the Cochran--Armitage trend test, yielding $Z=4.67$ and $p=3.07\times10^{-6}$.

Inter-rater reliability is reported with nominal Krippendorff's $\alpha$. The observed value is $\alpha=0.65$, with a bootstrap 95\% confidence interval of $[0.54,0.75]$. Pooled pairwise agreement is 84.0\%, and 62 of the 90 scenes (68.9\%) receive fully unanimous judgments.

\ifdefined\StandaloneSupp\else
\end{document}
\fi

\end{document}